\documentclass[10pt,twocolumn,letterpaper]{article}

\usepackage{titling}
\usepackage{iccv}
\usepackage{microtype}
\usepackage{times}
\usepackage{epsfig}
\usepackage{booktabs}
\usepackage{graphicx}
\usepackage{amsmath}
\usepackage{bm}
\usepackage{amssymb}
\usepackage{makecell}
\usepackage{xspace}
\usepackage{soul}
\usepackage{comment}
\usepackage[T1]{fontenc}

\newcommand{\rw}[1]{{\color{red}\{\textbf{Rundi:} #1\}}}

\newcommand{\cx}[1]{{\color{blue}\{\textbf{Chang:} #1\}}}
\newcommand{\rev}[1]{{\color{blue}#1}} 

\newcommand{\DC}{DeepCAD\xspace}

\newcommand{\cmdSOL}{\langle\mathtt{SOL}\rangle}
\newcommand{\cmdLine}{\mathtt{L}}
\newcommand{\cmdArc}{\mathtt{A}}
\newcommand{\cmdCirc}{\mathtt{R}}
\newcommand{\cmdExt}{\mathtt{E}}
\newcommand{\cmdEOS}{\langle\mathtt{EOS}\rangle}
\newcommand{\cmdNum}{N_c}
\newcommand{\paramNum}{N_P}
\newcommand{\seqCmd}{\{C_i\}_{i=1}^{\cmdNum}}
\newcommand{\seqCmdParam}{\{(c_i, P_i)\}_{i=1}^{\cmdNum}}
\newcommand{\seqCmdHat}{\{\hat C_i\}_{i=1}^{\cmdNum}}
\newcommand{\seqCmdParamHat}{\{(\hat c_i, \hat P_i)\}_{i=1}^{\cmdNum}}

\newcommand{\embDim}{d_\txt{E}}
\newcommand{\embCmd}{\bm{e}_i^{\text{cmd}}}
\newcommand{\embPm}{\bm{e}_i^{\text{param}}}
\newcommand{\embPos}{\bm{e}_i^{\text{pos}}}
\newcommand{\matCmd}{\mathsf{W}_{\text{cmd}}}
\newcommand{\matPmA}{\mathsf{W}_{\text{param}}^a}
\newcommand{\matPmB}{\mathsf{W}_{\text{param}}^b}
\newcommand{\matPos}{\mathsf{W}_{\text{pos}}}

\newcommand{\loss}{\mathcal{L}}
\newcommand{\ce}{\ell}

\newcommand{\figref}[1]{Fig.~\ref{fig:#1}}
\newcommand{\tabref}[1]{Table~\ref{tab:#1}}
\newcommand{\secref}[1]{Sec.~\ref{sec:#1}}
\newcommand{\eq}[1]{\eqref{eq:#1}}

\newcommand{\nameRel}{$\mathtt{Alt}\text{-}\mathtt{Rel}$ }
\newcommand{\nameTrans}{$\mathtt{Alt}\text{-}\mathtt{Trans}$ }
\newcommand{\nameArc}{$\mathtt{Alt}\text{-}\mathtt{ArcMid}$ }
\newcommand{\nameRegr}{$\mathtt{Alt}\text{-}\mathtt{Regr}$ }
\newcommand{\nameAug}{$\mathtt{Ours}\text{+}\mathtt{Aug}$ }

\newcommand{\txt}[1]{\textrm{#1}}

\newcommand{\accCmd}{\text{ACC}_\text{cmd}}
\newcommand{\accParam}{\text{ACC}_\text{param}}

\newcommand{\paraspace}{\vspace{-4.5mm}}
\newcommand{\subsecspace}{\vspace{-2.5mm}}

\usepackage[pagebackref=true,breaklinks=true,letterpaper=true,colorlinks,bookmarks=false]{hyperref}

\iccvfinalcopy 

\def\iccvPaperID{6743} 

\ificcvfinal\fi

\begin{document}

\title{DeepCAD: A Deep Generative Network for Computer-Aided Design Models}

\author{\qquad Rundi Wu \qquad Chang Xiao \qquad Changxi Zheng\\
Columbia University\\
{\tt\small \{rundi, chang, cxz\}@cs.columbia.edu}
}

\maketitle
\ificcvfinal\thispagestyle{empty}\fi

\begin{abstract}
\vspace{-2mm}
Deep generative models of 3D shapes have received a great deal of research
interest. Yet, almost all of them generate discrete shape representations, such as voxels,
point clouds, and polygon meshes. 
We present the first 3D generative model for a drastically different shape 
representation---describing a shape as a sequence of computer-aided design (CAD) operations.
Unlike meshes and point clouds, CAD models encode the user creation process of 3D shapes,
widely used in numerous industrial and engineering design tasks.
However, the sequential and irregular structure of CAD operations poses significant challenges 
for existing 3D generative models.
Drawing an analogy between CAD operations and natural language, 
we propose a CAD generative network based on the Transformer.
We demonstrate the performance of our
model for both shape autoencoding and random shape generation.
To train our network, we create a new CAD dataset
consisting of 178,238 models and their CAD construction sequences. 
We have made this dataset publicly available 
to promote future research on this topic.
\end{abstract}
\vspace{-4mm}

\vspace{-2mm}
\section{Introduction}
It is our human nature to imagine and invent, and to express our invention in
3D shapes.  This is what the paper and pencil were used for when Leonardo da
Vinci sketched his mechanisms; this is why such drawing tools as the parallel
bar, the French curve, and the divider were devised; and this is wherefore, in today's digital era,
the computer aided design (CAD) software have been used 
for 3D shape creation in a myriad of industrial sectors,
ranging from automotive and aerospace to
manufacturing and architectural design.

\begin{figure}[t!]
\centering
\includegraphics[width=0.99\linewidth, height=0.9\linewidth]{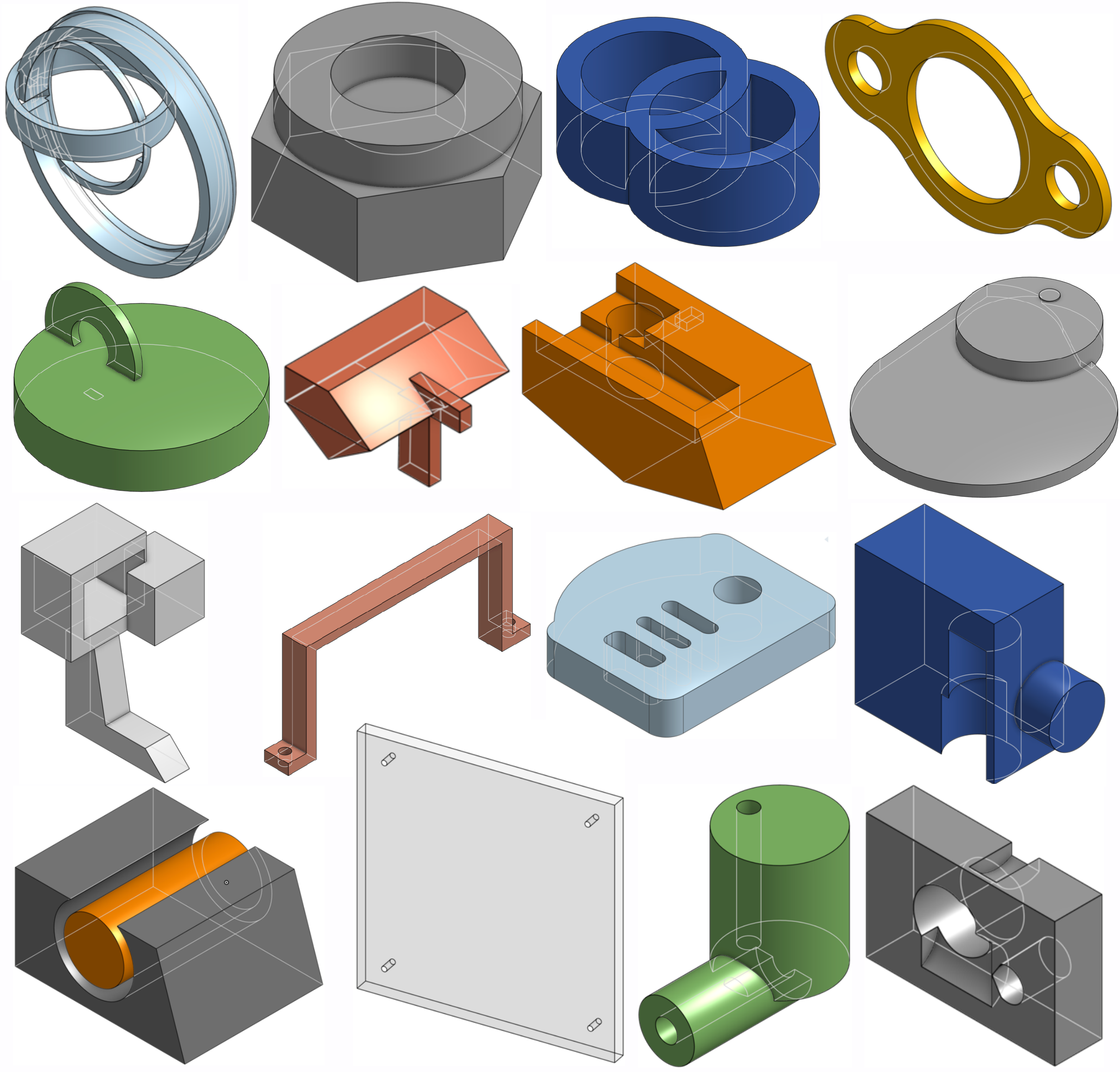}
\vspace{-0.5mm}
\caption{\textbf{A gallery of generated CAD designs.} Our generative network is able to
produce a diverse range of CAD designs. 
Each CAD model consists of a sequence of CAD operations with specific parameters.
The resulting 3D shapes are clean, have sharp geometric features, and can be readily user-edited.}
\vspace{-4mm}
\label{fig:teaser}
\end{figure}

Can the machine also invent 3D shapes?
Leveraging the striking advance in generative models of deep learning,
lots of recent research efforts have been directed to the generation of 3D models.
However, existing 3D generative models merely create 
computer discretization of 3D shapes:
3D point clouds~\cite{pmlr-v80-achlioptas18a,pointflow_yang2019,foldingnet_yang2018,ShapeGF,mo2019structurenet}, polygon meshes~\cite{altasnet_groueix2018,pixel2mesh_wang2018,polygen_nash20}, and levelset fields~\cite{imnet_chen2019,deepsdf_park2019,occnet_mescheder2019,pqnet_wu2020,bspnet_chen2020}.
Still missing is the ability to generate the very nature of 3D shape design\textemdash the drawing process.

We propose a deep generative network that outputs a sequence of operations
used in CAD tools (such as SolidWorks and AutoCAD) to construct a 3D shape. 
Generally referred as a \emph{CAD model},  
such an operational sequence represents the ``drawing'' process of shape creation. 
Today, almost all the industrial 3D designs start with CAD models.
Only until later in the production pipeline, if needed, they are discretized
into polygon meshes or point clouds.


To our knowledge, this is the first work toward a generative model of CAD
designs.  The challenge lies in the CAD design's sequential and parametric nature. 
A CAD model consists of a series of geometric operations (e.g., curve
sketch, extrusion, fillet, boolean, chamfer), each controlled by
certain parameters.  Some of the parameters are discrete options; others have
continuous values (more discussion in \secref{cad_rep}). These irregularities
emerge from the user creation process of 3D shapes, and thus contrast starkly
to the discrete 3D representations (i.e., voxels, point clouds, and meshes)
used in existing generative models.
In consequence, previously developed 3D generative models are unsuited for CAD model generation.

\paraspace
\paragraph{Technical contributions.}
To overcome these challenges, 
we seek a representation that reconciles the irregularities
in CAD models. We consider the most frequently used CAD operations (or commands), and unify them
in a common structure that encodes their command types, parameters, and sequential orders.
Next, drawing an analogy between CAD command sequences and natural languages,
we propose an autoencoder based on the Transformer network~\cite{transformer_NIPS2017}.
It embeds CAD models
into a latent space, and later decode a latent vector into a CAD command sequence.
To train our autoencoder, 
we further create a new dataset of CAD command sequences, one that is orders of magnitude larger than
the existing dataset of the same type.
We have also made this dataset publicly available\footnote{Code and data are available \href{https://github.com/ChrisWu1997/DeepCAD}{here}.} 
to promote future research on learning-based CAD designs.

Our method is able to generate plausible and diverse CAD designs (see \figref{teaser}).
We carefully evaluate its generation quality through a series of ablation studies.
Lastly, we end our presentation with an outlook on useful applications enabled by our CAD autoencoder.




\section{Related work}
\paragraph{Parametric shape inference.} 
Advance in deep learning has enabled neural network models that analyze 
geometric data and infer parametric shapes.
ParSeNet~\cite{parsenet_sharma2020} decomposes a 3D point cloud into 
a set of parametric surface patches.
PIE-NET~\cite{pienet_NEURIPS2020} extracts parametric boundary curves from 3D point clouds. 
UV-Net~\cite{uvnet_jayaraman2020} and BrepNet~\cite{lambourne2021brepnet} focus on encoding a parametric model's boundary curves and surfaces.
Li et al.~\cite{Sketch2CAD_2020_li} 
trained a neural network on synthetic data 
to convert 2D user sketches into CAD operations.
Recently, Xu et al.~\cite{xu2021inferring} applied neural-guided search to infer CAD modeling sequence from parametric solid shapes.

\paraspace
\paragraph{Generative models of 3D shapes.} 
Recent years have also witnessed increasing research interests on deep generative models 
for 3D shapes. Most existing methods generate 3D shapes in \emph{discrete} forms,
such as voxelized shapes~\cite{3dgan_wu2016,girdhar2016learning,deepmc_liao2018,GRASS_sig17li}, point
clouds~\cite{pmlr-v80-achlioptas18a,pointflow_yang2019,foldingnet_yang2018,ShapeGF,mo2019structurenet},
polygon meshes~\cite{altasnet_groueix2018,pixel2mesh_wang2018,polygen_nash20},
and {implicit signed distance fields}~\cite{imnet_chen2019,deepsdf_park2019,occnet_mescheder2019,pqnet_wu2020,bspnet_chen2020}.
The resulting shapes may still suffer from noise, lack sharp geometric features, and are not directly user editable.


Therefore, more recent works have sought neural network models that generate 3D shape
as a series of geometric operations. 
CSGNet~\cite{csgnet_sharma2018} 
infers a sequence of Constructive Solid Geometry (CSG)
operations based on voxelized shape input; and 
UCSG-Net~\cite{ucsg_kania2020} further advances the inference 
with no supervision from ground truth CSG trees.
Other than using CSG operations, several works synthesize 3D shapes using
their proposed domain specific
languages (DSLs)~\cite{tian2019learning,walke2020learning,mo2019structurenet,shapeAssembly_jones2020}.
For example, 
Jones et al.~\cite{shapeAssembly_jones2020} proposed
ShapeAssembly, a DSL that constructs 3D shapes by structuring 
cuboid proxies in a hierarchical and symmetrical fashion, and this structure can be 
generated through a variational autoencoder.





In contrast to all these works, our autoencoder network outputs CAD models
specified as a sequence of CAD operations. 
CAD models have become the standard shape representation 
in almost every sectors of industrial production.
Thus, the output from our network can be readily imported
into any CAD tools~\cite{web_autocad,web_fusion360,onshape_web} for user editing. It
can also be directly converted into other shape formats such as point clouds and polygon meshes.
To our knowledge, this is the first generative model directly producing CAD designs.


\paraspace
\paragraph{Transformer-based models.}
Technically, our work is related to the Transformer
network~\cite{transformer_NIPS2017}, which was introduced as an attention-based building
block for many natural language processing tasks~\cite{bert_devlin2018}.
The success of the Transformer network has also inspired its use in image processing
tasks~\cite{imageTrans_parmar2018,detr_carion2020,dosovitskiy2020image} 
and for other types of data~\cite{polygen_nash20,deepsvg_carlier2020,sceneformer_wang2020}.
Concurrent works~\cite{willis2021engineering,para2021sketchgen,ganin2021computer} on constrained CAD sketches generation also rely on
Transformer network. 


Also related to our work is DeepSVG~\cite{deepsvg_carlier2020}, a Transformer-based network
for the generation of Scalable Vector Graphic (SVG) images. SVG images are described 
by a collection of parametric primitives (such as lines and curves). Apart from limited in 2D, those primitives
are grouped with no specific order or dependence. In contrast, CAD commands are described in 3D;
they can be interdependent (e.g., through CSG boolean operations) and must follow a specific order.
We therefore seek a new way to encode CAD commands and their sequential order in a Transformer-based
autoencoder.


\vspace{-0.5mm}
\section{Method}\label{sec:method}
\vspace{-0.5mm}
We now present our \DC model, which revolves around a new representation of CAD
command sequences (\secref{rep}). Our CAD representation is specifically
tailored, for feeding into neural networks such as the proposed
Transformer-based autoencoder (\secref{autoencoder}). It also leads to a
natural objective function for training (\secref{training}).  To train our
network, we create a new dataset, one that is significantly larger than
existing datasets of the same type (\secref{dataset}), and one that itself can
serve beyond this work for future research.


\subsection{CAD Representation for Neural Networks}\label{sec:cad_rep}
\begin{figure}[t]
\centering
\includegraphics[width=0.99\linewidth, height=0.70\linewidth]{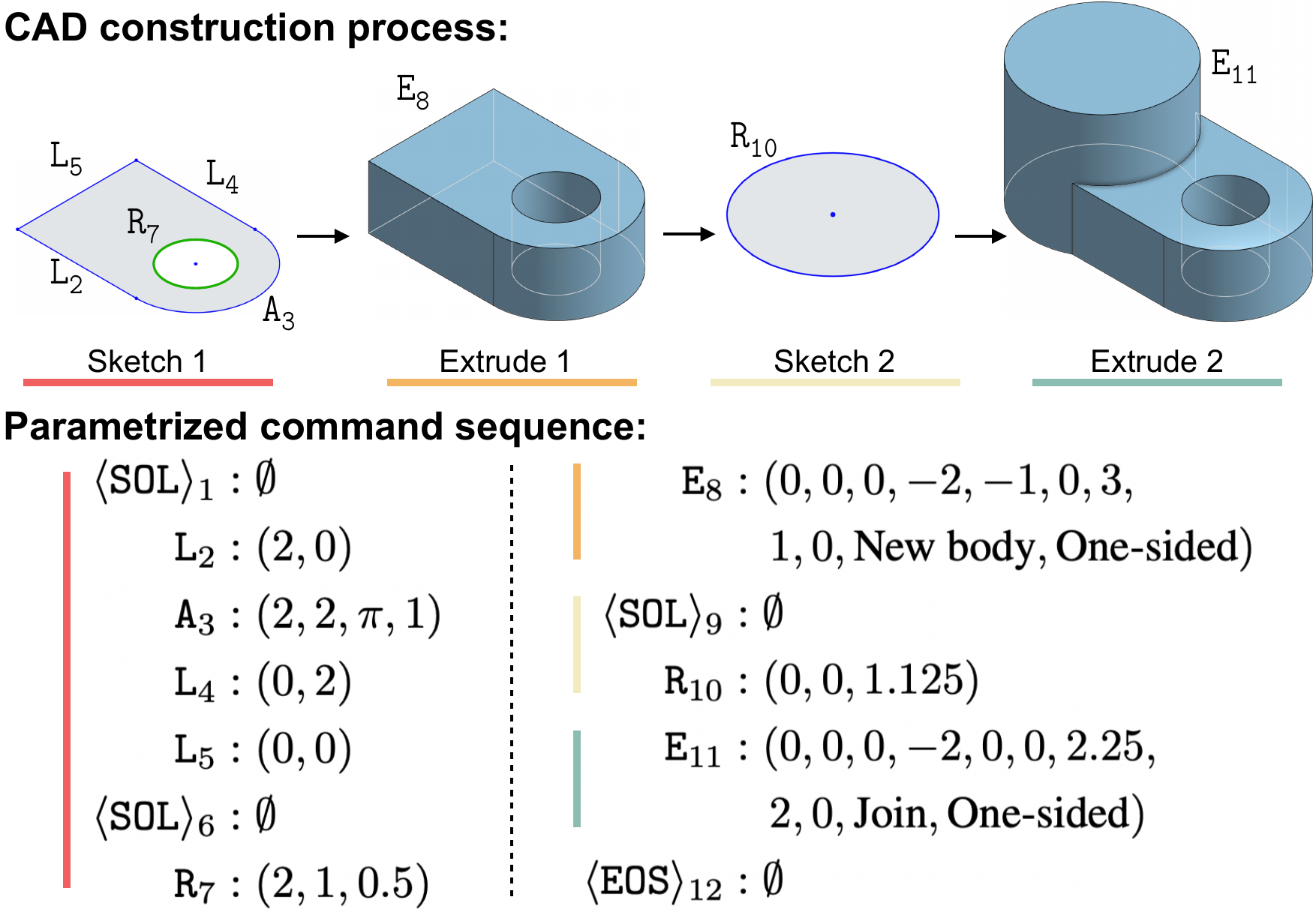}
\vspace{-0.5mm}
\caption{\textbf{A CAD model example} specified by the commands in \tabref{data_rep}.
(Top) the CAD model's
construction sequence, annotated with the command types. (Bottom) the 
command sequence description of the model. 
Parameter normalization and quantization are not shown in this case. 
In ``Sketch 1'', $\mathtt{L_2}$-$\mathtt{A_3}$-$\mathtt{L_4}$-$\mathtt{L_5}$ forms a loop
(in blue) and $\mathtt{C_7}$ forms another loop (in green), and the two loops bounds a 
sketch profile (in gray).}
\label{fig:data_rep}
\vspace{-3.5mm}
\end{figure}

\textbf{The CAD model} offers two levels of representation.
At the user-interaction level, a CAD model is described as a sequence of
operations that the user performs (in CAD software) to create a 
solid shape---for example, a user may $\mathtt{sketch}$ a closed 
curve profile on a 2D plane, and then $\mathtt{extrude}$ it into a 3D solid shape,
which is further processed by other operations such as a boolean 
$\mathtt{union}$ with another already created solid shape 
(see \figref{data_rep}). We refer to such a specification
as a CAD \emph{command sequence}.

Behind the command sequence is the CAD model's kernel representation,
widely known as the \emph{boundary representation} (or B-rep)~\cite{edge_1985Weiler,topologicalSF_Weiler1986}.
Provided a command sequence, its B-rep is automatically computed
(often through the industry standard library Parasolid). It consists of
topological components (i.e., vertices, parametric edges and faces) and the connections
between them to form a solid shape.

In this work, we aim for a generative model of CAD command sequences, not
B-reps.  This is because the B-rep is an abstraction from the command
sequence: a command sequence can be easily converted into a B-rep, but the
converse is hard, as different command sequences may result in the same B-rep.
Moreover, a command sequence is human-interpretable; it can be readily edited
(e.g., by importing them into CAD tools such as AutoCAD and Onshape),
allowing them to be used in various downstream applications.

\subsecspace
\vspace{-1mm}
\subsubsection{Specification of CAD Commands}
\vspace{-1mm}
Full-fledged CAD tools support a rich set of commands, although in practice only a small
fraction of them are commonly used. Here, we consider a subset of the
commands that are of frequent use (see \tabref{data_rep}).
These commands fall into two categories, namely \emph{sketch} and \emph{extrusion}.
While conceptually simple, they are sufficiently expressive to generate a wide variety of shapes, as has 
been demonstrated in~\cite{fusion2020willis}.

\begin{table}[t]
\centering
\resizebox{\columnwidth}{!}{%
\begin{tabular}{ccc}
\toprule
 \textbf{Commands} & \textbf{Parameters} \\ \toprule
$\cmdSOL$ & $\emptyset$ \\ \hline
 \makecell{$\cmdLine$ \\ (Line)} & \makecell[l]{
  $\begin{aligned} 
 \quad \quad\; x, y: \text{line end-point}
 \end{aligned}$
 }\\ \hline
 \makecell{$\cmdArc$ \\ (Arc)} & \makecell[l]{
 $\begin{aligned}
 \quad \quad\; x, y &: \text{arc end-point} \\ \alpha &: \text{sweep angle} \\ f &: \text{counter-clockwise flag}
 \end{aligned}$
 }\\ \hline
 \makecell{$\cmdCirc$ \\ (Circle)} & \makecell[l]{
 $\begin{aligned}
 \quad \quad\; x, y &: \text{center}\\ r &: \text{radius}
 \end{aligned}$
 }\\ \hline
 \makecell{$\cmdExt$ \\ (Extrude)} &  \makecell[c]{
 $\begin{aligned}
 \theta, \phi, \gamma &: \text{sketch plane orientation}\\ p_x, p_y, p_z &: \text{sketch plane origin} \\ s &: \text{scale of associated sketch profile} \\ e_1, e_2 &: \text{extrude distances toward both sides}\\ b &: \text{boolean type},\quad u : \text{extrude type}
 \end{aligned}$
 }\\ \hline
$\cmdEOS$ & $\emptyset$  \\ \bottomrule
\end{tabular}}
\caption{\textbf{CAD commands} and their parameters. $\cmdSOL$ indicates the start
    of a loop; $\cmdEOS$ indicates the end of the whole sequence.}\label{tab:data_rep}
\vspace{-3.5mm}
\end{table}

\paraspace
\paragraph{Sketch.}
Sketch commands are used to specify closed curves on a 2D plane in 3D space.
In CAD terminology, each closed curve is referred as a \emph{loop}, and one or more loops
form a closed region called a \emph{profile} (see ``Sketch 1'' in~\figref{data_rep}).
In our representation, a profile is described by a list of loops on its boundary;
a loop always starts with an indicator command $\cmdSOL$ followed by 
a series of curve commands $C_i$. 
We list all the curves on the loop in counter-clockwise order, beginning with the curve
whose starting point is at the most bottom-left; and the loops in a profile are sorted
according to the bottom-left corners of their bounding boxes.
Figure~\ref{fig:data_rep} illustrates two sketch profiles.

In practice, we consider three kinds of curve commands that are the most widely used:
draw a $\mathtt{line}$, an $\mathtt{arc}$, and a $\mathtt{circle}$. While other curve commands can be easily 
added (see~\secref{conclusion}), 
statistics from our large-scale real-world dataset (described in \secref{dataset})
show that these three types of commands constitute 92\% of the cases. 

Each curve command $C_i$ is described by its curve type $t_i\in\{\cmdSOL, \cmdLine, \cmdArc, \cmdCirc\}$
and its parameters listed in \tabref{data_rep}.
Curve parameters specify the curve's 2D location in the sketch plane's local frame of reference, whose own position and orientation
in 3D will be described shortly in the associated extrusion command.
Since the curves in each loop are concatenated one after another, 
for the sake of compactness we exclude the curve's starting position
from its parameter list; 
each curve always starts from the ending point of its predecessor in the loop.
The first curve always starts from the origin of the sketch plane, and the world-space coordinate of the origin 
is specified in the extrusion command.

In short, a sketch profile $S$ is described by a list of loops $S=[Q_1,\dots,Q_N]$, where each loop 
$Q_i$ consists of a series of curves starting from the indicator command $\cmdSOL$ 
(i.e., $Q_i=[\cmdSOL, C_1,\dots,C_{n_i}]$), and each curve command $C_j = (t_j, \bm{p}_j)$
specifies the curve type $t_i$ and its shape parameters $\bm{p}_j$ (see \figref{data_rep}).

\paraspace
\paragraph{Extrusion.}
The extrusion command serves two purposes. \textbf{1)} It extrudes a sketch profile from a 2D plane into
a 3D body, and the extrusion type can be either \emph{one-sided}, \emph{symmetric}, or \emph{two-sided} with respect to 
the profile's sketch plane.
\textbf{2)}~The command also specifies (through the parameter $b$ in \tabref{data_rep})
how to merge the newly extruded 3D body with the previously created
shape by one of the boolean operations: either creating a
\emph{new} body, or \emph{joining}, \emph{cutting} or \emph{intersecting} with
the existing body.

The extruded profile\textemdash which consists of one or more curve commands\textemdash 
is always referred to the one described immediately before the extrusion command.
The extrusion command therefore needs to define the 3D orientation of that profile's sketch plane and its 2D local
frame of reference. This is defined by a rotational matrix, determined by $(\theta,\gamma, \phi)$ parameters in \tabref{data_rep}. 
This matrix is to align the world frame of reference to the plane's local frame of reference, and to align $z$-axis 
to the plane's normal direction.
In addition, the command parameters include a scale factor $s$ of the extruded
profile; the rationale behind this scale factor will be discussed in \secref{rep}.

With these commands, we describe a CAD model $M$ as a sequence of curve commands interleaved
with extrusion commands (see \figref{data_rep}). 
In other words, $M$ is a command sequence $M=[C_1, \dots, C_{N_c}]$, where each $C_i$ has the form
$(t_i,\bm{p}_i)$ specifying the command type $t_i$ and parameters $\bm{p}_i$. 

\subsecspace
\subsubsection{Network-friendly Representation}\label{sec:rep}
\vspace{-1mm}
Our specification of a CAD model $M$ is akin to natural language.  The vocabulary
consists of individual CAD commands expressed sequentially to form
sentences. The subject of a sentence is the sketch profile; the predicate
is the extrusion.
This analogy suggests that we may leverage the network structures, such as the
Transformer network~\cite{transformer_NIPS2017}, succeeded in natural language
processing to fulfill our goal.

However, the CAD commands also differ from natural language in several aspects.
Each command has a different number of parameters.
In some commands (e.g., the extrusion), the parameters are a mixture of both continuous and discrete values,
and the parameter values span over different ranges (recall \tabref{data_rep}). These traits render the command sequences ill-posed for
direct use in neural networks.

To overcome this challenge, we regularize the dimensions of command sequences.
First, for each command, its parameters are stacked
into a 16$\times$1 vector, whose elements correspond to the collective parameters of all commands in \tabref{data_rep}
(i.e., $\bm{p}_i = [x, y, \alpha, f, r,\theta, \phi, \gamma, p_x, p_y, p_z, s, e_1, e_2, b, u]$). Unused parameters
for each command are simply set to be $-1$.
Next, we fix the total number $N_c$ of commands in every CAD model $M$. 
This is done by padding the CAD model's command sequence with the empty command $\cmdEOS$ until the sequence length reaches $N_c$.
In practice, we choose $N_c=60$, the maximal command sequence length appeared in our training dataset.

Furthermore, we unify continuous and discrete parameters by quantizing the continuous parameters.
To this end, we normalize every CAD model within a $2\times2\times2$ cube; 
we also normalize every sketch profile within its bounding box, and include a scale factor $s$ (in extrusion command) 
to restore the normalized profile into its original size. 
The normalization restricts the ranges of continuous parameters,
allowing us to quantize their values into 256 levels and express them using 8-bit integers.
As a result, all the command parameters possess only discrete sets of values. 

Not simply is the parameter quantization a follow-up of the common
practice for training {Transformer}-based
networks~\cite{PixeCNN_Salimans2017,polygen_nash20,sceneformer_wang2020}. 
Particularly for CAD models, it is crucial for improving the generation quality (as we empirically confirm in~\secref{results_ae}).
In CAD designs, certain geometric relations\textemdash such as parallel and perpendicular
 sketch lines\textemdash must be respected. However, if a generative model directly generates
continuous parameters, 
their values, obtained through parameter regression, are
prone to errors that will break these strict relations.
Instead, parameter quantization allows the network to ``classify''
parameters into specific levels, and thereby better respect learned geometric relations.

In \secref{results_ae}, we will present ablation studies that empirically justify our choices of CAD command representation.

\begin{figure}[t]
\centering
\includegraphics[width=0.9\linewidth,height=0.45\linewidth]{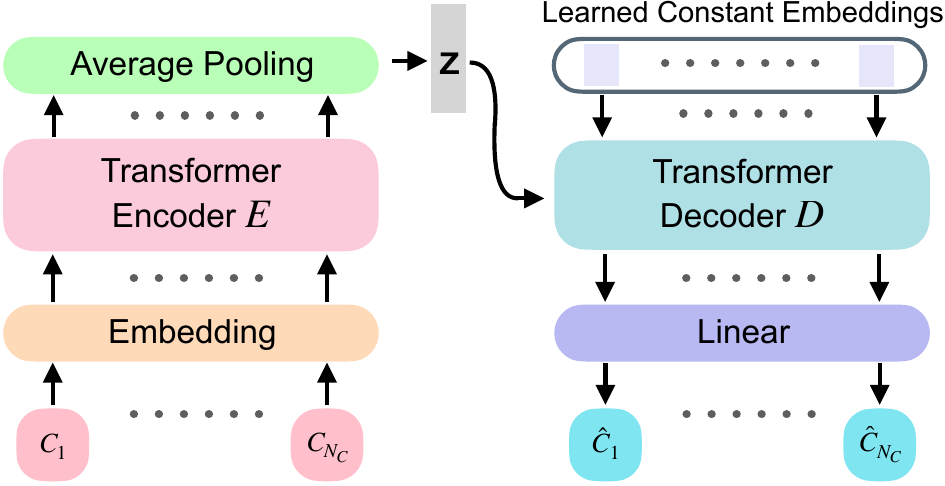}
\vspace{-0.5mm}
\caption{\textbf{Our network architecture.} The input CAD model, represented as
a command sequence $M=\seqCmd$ is first projected to an embedding space and then fed to the
encoder $E$ resulting in a latent vector $\bm{z}$. The decoder $D$ takes learned
constant embeddings as input, and also attends to the latent vector $\bm{z}$.
It then outputs the predicted command sequence $\hat M=\seqCmdHat$..}
\vspace{-3mm}
\label{fig:network}
\end{figure}

\subsection{Autoencoder for CAD Models}\label{sec:autoencoder}
We now introduce an autoencoder network that leverages our representation of CAD commands. 
Figure~\ref{fig:network} illustrates its structure, and more details are provided in \secref{supp_arch}
of supplementary document.
Once trained, the decoder part of the network will serve naturally as a CAD generative model.

Our autoencoder is based on the Transformer network, inspired by its success for 
processing sequential data~\cite{transformer_NIPS2017,bert_devlin2018,L_scher_2019}.
Our autoencoder takes as input a CAD command sequence $M=[C_1,\cdots,C_{N_c}]$, where $N_c$
is a fixed number (recall \secref{rep}). First, each command $C_i$ is projected separately onto a
continuous embedding space of dimension $\embDim=256$. Then, all the embeddings are put together to 
feed into an encoder $E$, which in turn outputs a latent vector $\bm{z}\in\mathbb{R}^{256}$.
The decoder takes the latent vector $\bm{z}$ as input, and outputs a generated
CAD command sequence $\hat{M}$.

\paraspace
\paragraph{Embedding.}
Similar in spirit to the approach in natural language
processing~\cite{transformer_NIPS2017}, we first project every command $C_i$
onto a common embedding space.  Yet, different from words in natural languages, a
CAD command $C_i=(t_i,\bm{p}_i)$ has two distinct parts: its command type $t_i$ and
parameters $\bm{p}_i$. We therefore formulate a different way of computing the embedding of $C_i$:
take it as a sum of three embeddings, that is, $\bm{e}(C_i) = \embCmd+\embPm+\embPos\in\mathbb{R}^{\embDim}$. 

The first embedding $\embCmd$ accounts for the command type $t_i$, given by 
$\embCmd=\matCmd\bm{\delta}_{i}^\txt{c}$. Here $\matCmd\in\mathbb{R}^{\embDim\times6}$  is a learnable matrix
and $\bm{\delta}_{i}^\txt{c}\in\mathbb{R}^6$ is a one-hot vector indicating the command type $t_i$ among the six command types. 

The second embedding $\embPm$ considers the command parameters. 
As introduced in \secref{rep}, every command has 16 parameters, each of which is quantized into an 8-bit integer.
We convert each of these integers
into a one-hot vector $\bm{\delta}_{i, j}^\txt{p}$ ($j=1..16$) of dimension $2^8+1=257$;
the additional dimension is to indicate that the parameter is unused in that command.
Stacking all the one-hot vectors into a matrix
$\bm{\delta}_{i}^\txt{p}\in\mathbb{R}^{257\times 16}$, we embed each parameter
separately using another learnable matrix
$\matPmB\in\mathbb{R}^{\embDim\times257}$, and then combine the individual
embeddings through a linear layer
$\matPmA\in\mathbb{R}^{\embDim\times 16\embDim}$, namely,
\begin{equation}
    \embPm=\matPmA\text{flat}(\matPmB\bm{\delta}_{i}^\txt{p}),
\end{equation}
where $\text{flat}(\cdot)$ flattens the input matrix to a vector.


Lastly, similar to~\cite{transformer_NIPS2017}, 
the positional embedding $\embPos$ is to indicate the index of the command $C_i$ in the whole 
command sequence, defined as $\embPos=\matPos\bm{\delta}_i$,
where $\matPos\in \mathbb{R}^{\embDim\times\cmdNum}$ is a learnable matrix and
$\bm{\delta}_i\in\mathbb{R}^{\cmdNum}$ is the one-hot vector filled with $1$ at
index $i$ and $0$ otherwise.

\paraspace
\paragraph{Encoder.} 
Our encoder $E$ is composed of four layers of Transformer blocks, each with
eight attention heads and feed-forward dimension of $512$.  The encoder takes
the embedding sequence $[\bm{e}_1,..,\bm{e}_{N_c}]$ as input, and outputs vectors
$[\bm{e}'_1,..,\bm{e}'_{N_c}]$; each has the same dimension $\embDim=256$. 
The output vectors are finally averaged
to produce a single $\embDim$-dimensional latent vector $\bm{z}$.

\paraspace
\paragraph{Decoder.} 
Also built on Transformer blocks, our decoder $D$ has the same
hyper-parameter settings as the encoder.  
It takes as input learned constant embeddings while also attending to the latent vector $\bm{z}$\textemdash
similar input structure has been used in~\cite{detr_carion2020,deepsvg_carlier2020}. 
Output from the last Transformer block is fed into a linear layer
to predict a CAD command sequence $\hat{M}=[\hat{C}_1,..,\hat{C}_{N_c}]$,
including both the command type $\hat{t}_i$ and parameters $\hat{\bm{p}}_i$ for each command.
As opposed to the autoregressive strategy commonly used in
natural language processing~\cite{transformer_NIPS2017}, we adopt the feed-forward
strategy~\cite{detr_carion2020,deepsvg_carlier2020}, and the prediction of our
model can be factorized as 
\begin{equation}
    p(\hat M|z,\Theta)=\overset{\cmdNum}{\underset{i=1}{\prod}}p(\hat t_i,\hat{\bm{p}}_i|z,\Theta),
\end{equation}
where $\Theta$ denotes network parameters of the decoder.


\subsection{Creation of CAD Dataset}\label{sec:dataset}

Several datasets of CAD designs exist, but none of them suffice for our training.
In particular, the ABC dataset~\cite{abc_2019_CVPR} collects about $1$ million CAD designs from Onshape,
a web-based CAD tool and repository~\cite{onshape_web}.
Although this is a large-scale dataset, its CAD designs are provided in B-rep format, with no sufficient information
to recover how the designs are constructed by CAD operations.
The recent \emph{Fusion 360 Gallery} dataset~\cite{fusion2020willis}
offers CAD designs constructed by profile sketches and extrusions, and it
provides the CAD command sequence for each design.
However, this dataset has only $\sim8000$ CAD designs,
not enough for training a well generalized generative model.

We therefore create a new dataset that is large-scale and provides CAD command sequences. 
Apart from using it to train our autoencoder network, this dataset may also serve for future research.
We have made it publicly available.

To create the dataset, we also leverage Onshape's CAD repository and its
developer API~\cite{onshape_dev} to parse the CAD designs. We start from the
ABC dataset. For each CAD model, the dataset provides a link to Onshape's original
CAD design.  We then use Onshape's domain specific language (called
FeatureScript~\cite{onshape_fs}) to parse CAD operations and parameters used in
that design.  For CAD models that use the operations beyond sketch and
extrusion, we simply discard them. For the rest of the models, we use a FeatureScript
program to extract the sketch profiles and extrusions, and express them using the commands listed in 
\tabref{data_rep}.

In the end, we collect a dataset with {178,238} CAD designs all described as CAD command sequences. This is orders of magnitude
larger than the existing dataset of the same type~\cite{fusion2020willis}.
The dataset is further split into training, validation and test sets by 90\%-5\%-5\% in a random fashion,
ready to use in training and testing. Figure~\ref{fig:dataset_gallery} in the supplementary document
samples some CAD models from our dataset.

\subsection{Training and Runtime Generation}\label{sec:training}
\paragraph{Training.}
Leveraging the dataset, we train our autoencoder network using the standard Cross-Entropy loss.
Formally, we define the loss between the predicted CAD model $\hat M$ and the ground truth model $M$ as 
\begin{equation}\label{eq:training}
    \loss=\overset{\cmdNum}{\underset{i=1}{\sum}}\ce(\hat t_i, t_i)+\beta\overset{\cmdNum}{\underset{i=1}{\sum}}\overset{N_P}{\underset{j=1}{\sum}}\ce(\hat{\bm{p}}_{i, j}, \bm{p}_{i, j}),
\end{equation}
where $\ce(\cdot,\cdot)$ denotes the standard Cross-Entropy, $N_p$ is the number of parameters ($N_p=16$ in our examples),
and $\beta$ is a weight to balance both terms ($\beta=2$ in our examples).
Note that in the ground-truth command sequence, some commands are empty (i.e., the padding command $\cmdEOS$) and some command parameters
are unused (i.e., labeled as $-1$). In those cases, their corresponding contributions to the summation terms in~\eq{training} 
are simply ignored.


The training process uses the Adam optimizer~\cite{kingma2014adam} with a learning rate $0.001$ and a linear warm-up period of $2000$ initial steps. 
We set a dropout rate of $0.1$ for all Transformer blocks and apply gradient clipping of $1.0$ in back-propagation. 
We train the network for $1000$ epochs with a batch size of $512$.

\paraspace
\paragraph{CAD generation.}
Once the autoencoder is well trained, we can represent a CAD model using a $256$-dimensional latent vector $\bm{z}$. 
For automatic generation of CAD models, we employ the
latent-GAN technique~\cite{pmlr-v80-achlioptas18a,imnet_chen2019,pqnet_wu2020}
on our learned latent space.  
The generator and discriminator are both as simple as a multilayer perceptron (MLP) network with four hidden layers,
and they are trained using Wasserstein-GAN training strategy with gradient penalty~\cite{wgan_pmlr-v70-arjovsky17a,wgangp_Gulrajani2017}.  
In the end, to generate a CAD model, we sample a random vector
from a multivariate Gaussian distribution and feeding it into the GAN's generator.
The output of the GAN is a latent vector $\bm{z}$ input to our Transformer-based decoder.

\section{Experiments}
In this section, we evaluate our autoencoder network from two perspectives:
the autoencoding of CAD models (\secref{results_ae}) and latent-space shape generation (\secref{results_gen}).
We also discuss possible applications that can benefit from our CAD generative model (\secref{results_dis}).

There exist no previous generative models for CAD designs, and thus no methods for our model to direct compare with.
Our goal here is to understand the performance of our model under different metrics, and justify
the algorithmic choices in our model through a series of ablation studies.

\begin{table}[t]
\begin{center}
\resizebox{\linewidth}{!}{
\begin{tabular}{lcccc}
\toprule
Method & \makecell{$\accCmd\uparrow$} & \makecell{$\accParam\uparrow$} & \makecell{median \\ CD$\downarrow$} & \makecell{Invalid \\ Ratio$\downarrow$}\\
\midrule

\nameAug    & \textbf{99.50} & \textbf{97.98} & \textbf{0.752} & \textbf{2.72} \\
$\mathtt{Ours}$ & 99.36 & 97.47 & 0.787 & 3.30 \\
\nameArc & 99.34 & 97.31 & 0.790 & 3.26 \\
\nameTrans  & 99.33 & 97.56 & 0.792 & 3.30 \\
\nameRel    & 99.33 & 97.66 & 0.863 & 3.51 \\ 
\nameRegr   &  -    & -     & 2.142 & 4.32 \\

\bottomrule
\end{tabular}}
\end{center}
\vspace{-2mm}
\caption{\textbf{Quantitative evaluation of autoencoding.} 
$\accCmd$ and $\accParam$ are both multiplied by $100\%$, and CD is multiplied by $10^3$. $\uparrow$: a higher metric value
indicates better autoencoding quality. $\downarrow$: a lower metric value is better. 
ACC values for \nameRegr are not available since \nameRegr does not use quantized parameters.
}
\label{tab:results_ae}
\vspace{-3mm}
\end{table}

\subsection{Autoencoding of CAD Models}\label{sec:results_ae}
The autoencoding performance has often been used to indicate the extent to which
the generative model can express the target data distribution~\cite{pmlr-v80-achlioptas18a,imnet_chen2019,altasnet_groueix2018}.
Here we use our autoencoder network to encode a CAD model $M$ absent from the training dataset; we then decode
the resulting latent vector into a CAD model $\hat{M}$. The autoencoder is evaluated by the difference between $M$ and $\hat{M}$.

\begin{figure}[t]
\centering
\includegraphics[width=0.99\linewidth,height=1.25\linewidth]{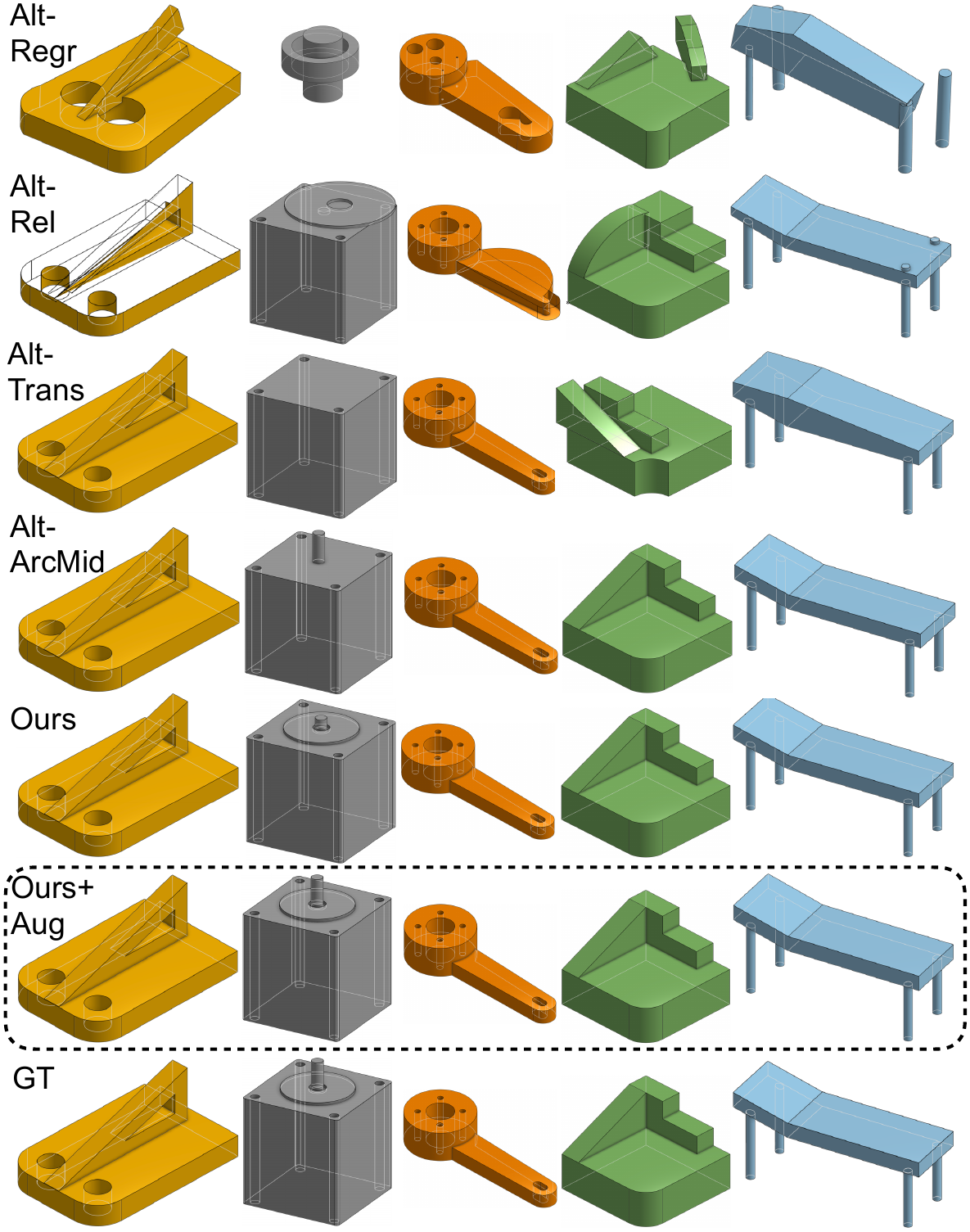}
\vspace{-1mm}
\caption{\textbf{Comparison of autoencoding results.} Hidden edges are
also rendered visible (white). Ground truth (GT) is shown in the bottom row.
Our best results are highlighted in the dash-line box.}\label{fig:results_ae}
\vspace{-3mm}
\end{figure}

\paraspace
\paragraph{Metrics.}
To thoroughly understand our autoencoder's performance, we measure the
difference between $M$ and $\hat{M}$ in terms of both the CAD commands 
and the resulting 3D geometry.
We propose to evaluate command accuracy using two metrics, namely
\emph{Command Accuracy} ($\accCmd$) and \emph{Parameter Accuracy} ($\accParam$).  
The former measures the correctness of the predicted CAD command type, defined as
\begin{equation}
    \accCmd = \frac{1}{\cmdNum}\sum_{i=1}^{\cmdNum}\mathbb{I}[t_i = \hat t_i].
\end{equation}
Here the notation follows those in \secref{method}. $\cmdNum$ denote the total number of CAD commands, 
and $t_i$ and $\hat t_i$ are the ground-truth and recovered command types, respectively.
$\mathbb{I}[\cdot]$ is the indicator function (0 or 1).


Once the command type is correctly recovered, we also evaluate the correctness of the command parameters.
This is what Parameter Accuracy ($\accParam$) is meant to measure:
\begin{equation}
    \accParam = \frac{1}{K} \sum_{i=1}^{\cmdNum} \sum_{j=1}^{|\hat{\bm{p}}_i|} \mathbb{I}[|\bm{p}_{i,j} - \hat{\bm{p}}_{i,j}|<\eta] \mathbb{I}[t_i = \hat t_i],
\end{equation}
where $K=\sum_{i=1}^{\cmdNum}\mathbb{I}[t_i = \hat t_i]|\bm{p}_i|$ is the total number of parameters in all correctly recovered commands.
Note that $\bm{p}_{i,j}$ and $\hat{\bm{p}}_{i,j}$ are both quantized into 8-bit integers.
$\eta$ is chosen as a tolerance threshold accounting for the parameter quantization. 
In practice, we use $\eta=3$ (out of 256 levels).

To measure the quality of recovered 3D geometry, we use \emph{Chamfer Distance}
(CD), the metric used in many previous generative models of
discretized shapes (such as point clouds)~\cite{pmlr-v80-achlioptas18a,altasnet_groueix2018,imnet_chen2019}.  Here, we evaluate CD
by uniformly sampling $2000$ points on the surfaces of reference
shape and recovered shape, respectively; and measure CD 
between the two sets of points.  Moreover, it is not guaranteed that the output
CAD command sequence always produces a valid 3D shape.  In rare cases, the
output commands may lead to an invalid topology, and thus no point
cloud can be extracted from that CAD model.  We therefore also report the
\emph{Invalid Ratio}, the percentage of the output CAD models that
fail to be converted to point clouds.
 
\paraspace
\paragraph{Comparison methods.} 
Due to the lack of existing CAD generative models,
we compare our model with several variants in order to justify
our data representation and training strategy. 
In particular, we consider the following variants.

\nameRel represents curve positions relative to the position of its predecessor
curve in the loop.  It contrasts to our model, which uses
absolute positions in curve specification. 

\nameTrans includes in the extrusion command the starting point position of the
loop (in addition to the origin of the sketch plane). Here the starting point
position and the plane's origin are in the world frame of reference of the CAD model.
In contrast, our proposed method includes only the sketch plane's origin, and 
the origin is translated to the loop's starting position\textemdash it is therefore more compact.


\nameArc specifies an arc using its ending and middle point positions, but not 
the sweeping angle and the counter-clockwise flag used in \tabref{data_rep}.

\nameRegr regresses all parameters of the CAD commands using the standard
mean-squared error in the loss function. 
Unlike the model we propose,
there is no need to quantize continuous parameters in this approach.

\nameAug uses the same data representation and training objective as our proposed
solution, but it augment the training dataset by including randomly composed CAD command
sequences (although the augmentation may be an invalid CAD sequence in few cases).

More details about these variants are described in \secref{supp_ae} of the supplementary document.

\begin{figure}[t]
    \centering
\includegraphics[width=0.98\linewidth, height=0.65\linewidth]{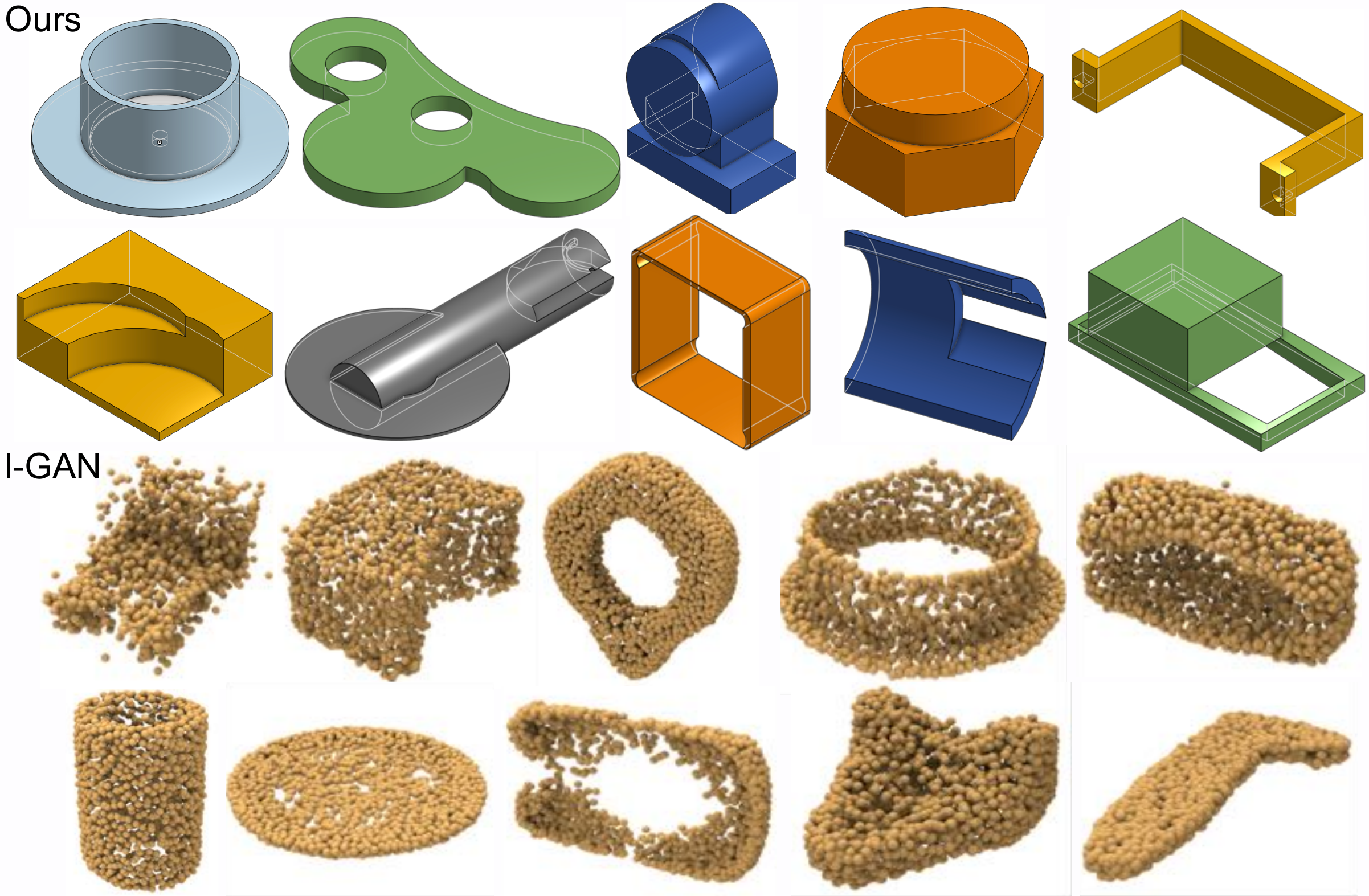}
\vspace{-1mm}
\caption{\textbf{Randomly generated 3D shapes} from our model (top) and l-GAN (bottom).
   }\label{fig:results_gen}
\vspace{-1.5mm}
\end{figure}
\begin{table}[t!]
\begin{center}
\begin{tabular}{lccc}
\toprule
                 Method & COV$\uparrow$ & MMD$\downarrow$ & JSD$\downarrow$ \\ \midrule
                 Ours & \textbf{78.13} & 1.45 & \textbf{3.76} \\
                 l-GAN & 77.73 & \textbf{1.27} & 5.02\\
\bottomrule
\end{tabular}
\end{center}
\vspace{-3mm}
\caption{\textbf{Shape generation measured under point-cloud metrics.} 
We use the metrics in l-GAN~\cite{pmlr-v80-achlioptas18a}.
Both MMD and JSD are multiplied by $10^2$. $\uparrow$: the higher the better,
$\downarrow$: the lower the better.}
\label{tab:results_gen}
\vspace{-4mm}
\end{table}

\paraspace
\paragraph{Discussion of results.} 
The quantitative results are report in \tabref{results_ae}, and more detailed CD scores are 
given in \tabref{supp_cd} of the supplementary document. In general,
\nameAug (i.e., training with synthetic data augmentation) achieves the best
performance, suggesting that randomly composed data can improve the
network's generalization ability.  
The performance of \nameArc is similar to $\mathtt{Ours}$. This means that 
middle-point representation is a viable alternative to represent arcs.
Moreover, \nameTrans performs slightly worse in terms of CD than $\mathtt{Ours}$ 
(e.g., see the green model in \figref{results_ae}). 

Perhaps more interestingly, 
while \nameRel has high parameter accuracy ($\accParam$), even higher than $\mathtt{Ours}$,
it has a relatively large CD score and sometimes invalid topology: for example, 
the yellow model in the second row of \figref{results_ae} has two triangle loops intersecting with
each other, resulting in invalid topology.
This is caused by the errors of the predicted curve positions. 
In \nameRel, curve positions are specified with respect to its predecessor curve, and thus the error accumulates
along the loop.


Lastly, \nameRegr, not quantizing continuous parameters, suffers from larger errors that 
may break curcial geometric relations such as parallel and perpendicular edges
(e.g., see the orange model in~\figref{results_ae}).

\paraspace
\paragraph{Cross-dataset generalization.}
We also verify the generalization of our autoencoder: 
we take our autoencoder trained on our created dataset and evaluate 
it on the smaller dataset provided in~\cite{fusion2020willis}. These datasets are constructed
from different sources: ours is based on models from Onshape repository, while theirs is produced
from designs in Autodesk Fusion 360. Nonetheless, our network generalizes well on their dataset,
achieving comparable quantitative performance (see \secref{supp_fusion} in supplementary document).


\subsection{Shape Generation} \label{sec:results_gen}
Next, we evaluate CAD model generation from latent vectors (described in \secref{training}).
Some examples of our generated CAD models are shown in~\figref{teaser},
and more results are presented in \figref{supp_gen} of the supplementary document.


Since there are no existing generative models for CAD designs, we choose to
compare our model with \emph{l-GAN}~\cite{pmlr-v80-achlioptas18a}, a widely
studied point-cloud 3D shape generative model.
We note that our goal is \emph{not} to show the superiority one over another, as the two generative 
models have different application areas.
Rather, we demonstrate our model's ability to generate comparable shape quality even under the
metrics for point-cloud generative models.
Further, shapes from our model, as shown in \figref{results_gen}, have much sharper geometric details, and
they can be easily user edited (\figref{results_edit}).




\begin{figure}[t]
    \centering
\includegraphics[width=0.97\linewidth, height=0.32\linewidth]{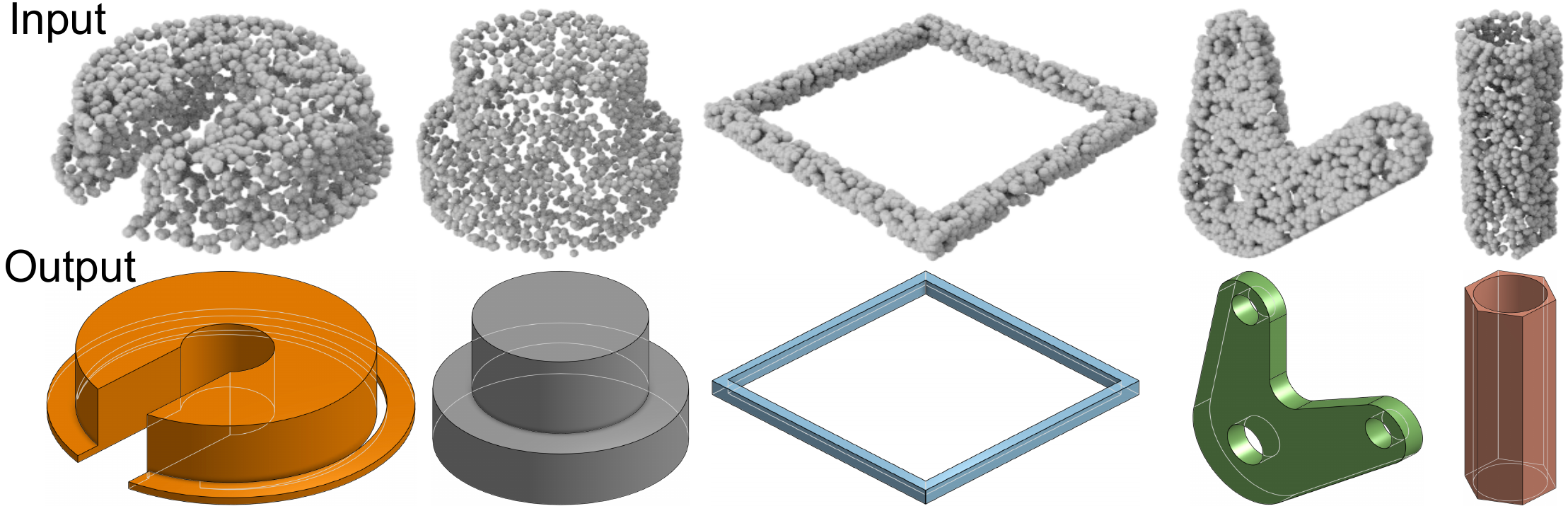}
\vspace{0mm}
\caption{\textbf{CAD model reconstruction from point clouds}. (Top) input point clouds.
   (Bottom) reconstructed CAD models.}
\label{fig:pc2cad}
\vspace{-5mm}
\end{figure}

\paraspace
\paragraph{Metrics.}
For quantitative comparison with point-cloud generative models, we follow the metrics
used in l-GAN~\cite{pmlr-v80-achlioptas18a}. 
Those metrics measure the discrepancy between two sets of 3D point-cloud shapes,
the set $\mathcal{S}$ of ground-truth shapes and the set $\mathcal{G}$ of generated shapes.
In particular,
\emph{Coverage (COV)} measures what percentage
of shapes in $\mathcal{S}$ can be well approximated by shapes in $\mathcal{G}$.
\emph{Minimum Matching Distance (MMD)} measures the fidelity of $\mathcal{G}$ 
through the minimum matching distance between two point clouds from $\mathcal{S}$ and $\mathcal{G}$.
\emph{Jensen-Shannon Divergence (JSD)} is the standard statistical distance, measuring the similarity between 
the point-cloud distributions of $\mathcal{S}$ and $\mathcal{G}$.
Details of computing these metrics are present in the supplement (\secref{supp_metrics}).

\paraspace
\paragraph{Discussion of results.}
Figure~\ref{fig:results_gen} illustrates some output examples from our CAD generative model 
and l-GAN.
We then convert ground-truth and generated CAD models into point clouds, 
and evaluate the metrics.  The results are reported in \tabref{results_gen},
indicating that our method has comparable performance as l-GAN in terms of
the point-cloud metrics. Nevertheless, CAD models, thanks to their parametric 
representation, have much smoother surfaces and sharper geometric features than point clouds.



\subsection{Future Applications}\label{sec:results_dis}
The CAD generative model can serve as a fundamental algorithmic block in many applications.
While our work focuses on the generative model itself, not the downstream applications,
here we discuss its use in two scenarios.

With the CAD generative model, one can take a point cloud (e.g., acquired through 3D scanning)
and reconstruct a CAD model. As a preliminary demonstration, we use our autoencoder to encode
a CAD model $M$ into a latent vector $\bm{c}$. We then leverage the PointNet++ encoder~\cite{pointnet2_qi2017},
training it to encode the point-cloud representation of $M$ into the same latent vector $\bm{c}$.
At inference time, provided a point cloud, we use PointNet++ encoder to map it into a latent vector,
followed by our auotoencocder to decode into a CAD model. 
We show some visual examples in \figref{pc2cad}
and quan-titative results in the supplementary document (\tabref{results_pc2cad}).

\begin{figure}[t!]
    \centering
\includegraphics[width=0.97\linewidth]{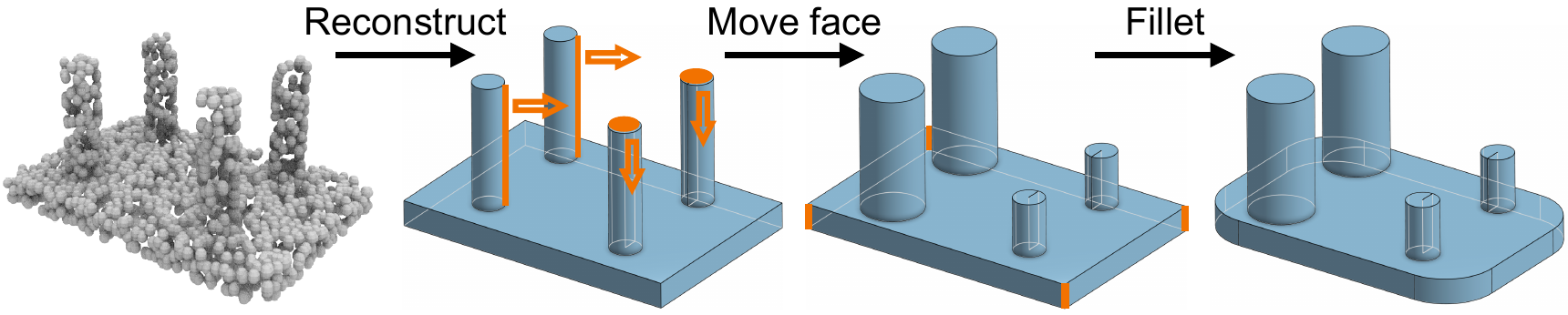}
\vspace{-0.5mm}
\caption{\textbf{User Editing.} Our reconstructed CAD model can be easily edited in any CAD tools.
   Here, the regions that undergo CAD operations are
   highlighted in orange color.}
\label{fig:results_edit}
\vspace{-5mm}
\end{figure}

 
Furthermore, the generated CAD model can be directly imported into CAD tools for user editing (see \figref{results_edit}). 
This is a unique feature enabled by the CAD generative model,
as the user editing on point clouds or polygon meshes would be much more troublesome.


\vspace{-0.9mm}
\section{Discussion and Conclusion}\label{sec:conclusion}
\vspace{-0.4mm}
Toward the CAD generative model, there are several limitations in our approach. 
At this point, we have considered three most widely used types of curve commands (line, arc, circle),
but other curve commands can be easily added as well. For example,
a cubic B\'ezier curve can be
specified by three control points together with the starting point from the
ending position of its predecessor. These parameters can be structured in the same way 
as described in \secref{cad_rep}.
Other operations, such as revolving a sketch, can be encoded in a way
similar to the extrusion command.
However, certain CAD operations such as \emph{fillet}
operate on parts of the shape boundary, and thus they require a reference
to the model's B-rep, not just other commands. To incorporate those commands in the generative model 
is left for future research.

Not every CAD command sequence can produce topologically valid shape. 
Our generative network cannot guarantee topological soundness of its output CAD sequences.
In practice, the generated CAD command sequence rarely fails. 
The failure becomes more likely as the command sequence becomes quite long.
We present and analyze some failure cases in \secref{supp_failure} of the supplementary document,
providing some fodder for future research.

In summary, we have presented DeepCAD, a deep generative model 
for CAD designs.
Almost all previous 3D generative models produce
discrete 3D shapes such as voxels, point clouds, and meshes.
This work, to our knowledge, is the first generative model for CAD designs.
To this end, we also introduce a large dataset of CAD models, each represented 
as a CAD command sequence.

\vspace{-3.5mm}
\paragraph{Acknowledgements.}
We thank the anonymous reviewers for their constructive feedback. This work was
partially supported by the National Science Foundation (1910839 and 1816041).

{\small
\bibliographystyle{ieee_fullname}
\bibliography{egbib}

\begin{thebibliography}{10}\itemsep=-1pt

\bibitem{web_autocad}
Autocad.
\newblock \url{https://www.autodesk.com/products/autocad}.

\bibitem{web_fusion360}
Fusion 360.
\newblock \url{https://www.autodesk.com/products/fusion-360}.

\bibitem{onshape_web}
Onshape.
\newblock \url{http://http://onshape.com}.

\bibitem{onshape_dev}
Onshape developer documentation.
\newblock \url{https://onshape-public.github.io/docs/}.

\bibitem{onshape_fs}
Onshape featurescript.
\newblock \url{https://cad.onshape.com/FsDoc/}.

\bibitem{pmlr-v80-achlioptas18a}
Panos Achlioptas, Olga Diamanti, Ioannis Mitliagkas, and Leonidas Guibas.
\newblock Learning representations and generative models for 3{D} point clouds.
\newblock In Jennifer Dy and Andreas Krause, editors, {\em Proceedings of the 35th International Conference on Machine Learning}, volume~80 of {\em Proceedings of Machine Learning Research}, pages 40--49, Stockholmsmässan, Stockholm Sweden, 10--15 Jul 2018. PMLR.

\bibitem{wgan_pmlr-v70-arjovsky17a}
Martin Arjovsky, Soumith Chintala, and L{\'e}on Bottou.
\newblock {W}asserstein generative adversarial networks.
\newblock In Doina Precup and Yee~Whye Teh, editors, {\em Proceedings of the 34th International Conference on Machine Learning}, volume~70 of {\em Proceedings of Machine Learning Research}, pages 214--223, International Convention Centre, Sydney, Australia, 06--11 Aug 2017. PMLR.

\bibitem{ShapeGF}
Ruojin Cai, Guandao Yang, Hadar Averbuch-Elor, Zekun Hao, Serge Belongie, Noah Snavely, and Bharath Hariharan.
\newblock Learning gradient fields for shape generation.
\newblock In {\em Proceedings of the European Conference on Computer Vision (ECCV)}, 2020.

\bibitem{detr_carion2020}
Nicolas Carion, Francisco Massa, Gabriel Synnaeve, Nicolas Usunier, Alexander Kirillov, and Sergey Zagoruyko.
\newblock End-to-end object detection with transformers.
\newblock In {\em European Conference on Computer Vision}, pages 213--229. Springer, 2020.

\bibitem{deepsvg_carlier2020}
Alexandre Carlier, Martin Danelljan, Alexandre Alahi, and Radu Timofte.
\newblock Deepsvg: A hierarchical generative network for vector graphics animation.
\newblock In H. Larochelle, M. Ranzato, R. Hadsell, M.~F. Balcan, and H. Lin, editors, {\em Advances in Neural Information Processing Systems}, volume~33, pages 16351--16361. Curran Associates, Inc., 2020.

\bibitem{bspnet_chen2020}
Zhiqin Chen, Andrea Tagliasacchi, and Hao Zhang.
\newblock Bsp-net: Generating compact meshes via binary space partitioning.
\newblock In {\em Proceedings of the IEEE/CVF Conference on Computer Vision and Pattern Recognition}, pages 45--54, 2020.

\bibitem{imnet_chen2019}
Zhiqin Chen and Hao Zhang.
\newblock Learning implicit fields for generative shape modeling.
\newblock In {\em Proceedings of the IEEE/CVF Conference on Computer Vision and Pattern Recognition}, pages 5939--5948, 2019.

\bibitem{bert_devlin2018}
Jacob Devlin, Ming-Wei Chang, Kenton Lee, and Kristina Toutanova.
\newblock Bert: Pre-training of deep bidirectional transformers for language understanding.
\newblock {\em arXiv preprint arXiv:1810.04805}, 2018.

\bibitem{dosovitskiy2020image}
Alexey Dosovitskiy, Lucas Beyer, Alexander Kolesnikov, Dirk Weissenborn, Xiaohua Zhai, Thomas Unterthiner, Mostafa Dehghani, Matthias Minderer, Georg Heigold, Sylvain Gelly, et~al.
\newblock An image is worth 16x16 words: Transformers for image recognition at scale.
\newblock {\em arXiv preprint arXiv:2010.11929}, 2020.

\bibitem{ganin2021computer}
Yaroslav Ganin, Sergey Bartunov, Yujia Li, Ethan Keller, and Stefano Saliceti.
\newblock Computer-aided design as language.
\newblock {\em arXiv preprint arXiv:2105.02769}, 2021.

\bibitem{girdhar2016learning}
Rohit Girdhar, David~F Fouhey, Mikel Rodriguez, and Abhinav Gupta.
\newblock Learning a predictable and generative vector representation for objects.
\newblock In {\em European Conference on Computer Vision}, pages 484--499. Springer, 2016.

\bibitem{altasnet_groueix2018}
Thibault Groueix, Matthew Fisher, Vladimir~G Kim, Bryan~C Russell, and Mathieu Aubry.
\newblock A papier-m{\^a}ch{\'e} approach to learning 3d surface generation.
\newblock pages 216--224, 2018.

\bibitem{wgangp_Gulrajani2017}
Ishaan Gulrajani, Faruk Ahmed, Martin Arjovsky, Vincent Dumoulin, and Aaron Courville.
\newblock Improved training of wasserstein gans.
\newblock In {\em Proceedings of the 31st International Conference on Neural Information Processing Systems}, NIPS'17, pages 5769--5779, USA, 2017. Curran Associates Inc.

\bibitem{uvnet_jayaraman2020}
Pradeep~Kumar Jayaraman, Aditya Sanghi, Joseph Lambourne, Thomas Davies, Hooman Shayani, and Nigel Morris.
\newblock Uv-net: Learning from curve-networks and solids.
\newblock {\em arXiv preprint arXiv:2006.10211}, 2020.

\bibitem{shapeAssembly_jones2020}
R.~Kenny Jones, Theresa Barton, Xianghao Xu, Kai Wang, Ellen Jiang, Paul Guerrero, Niloy~J. Mitra, and Daniel Ritchie.
\newblock Shapeassembly: Learning to generate programs for 3d shape structure synthesis.
\newblock {\em ACM Transactions on Graphics (TOG), Siggraph Asia 2020}, 39(6):Article 234, 2020.

\bibitem{ucsg_kania2020}
Kacper Kania, Maciej Zi{\k{e}}ba, and Tomasz Kajdanowicz.
\newblock Ucsg-net--unsupervised discovering of constructive solid geometry tree.
\newblock {\em arXiv preprint arXiv:2006.09102}, 2020.

\bibitem{kingma2014adam}
Diederik~P. Kingma and Jimmy Ba.
\newblock Adam: A method for stochastic optimization, 2014.

\bibitem{abc_2019_CVPR}
Sebastian Koch, Albert Matveev, Zhongshi Jiang, Francis Williams, Alexey Artemov, Evgeny Burnaev, Marc Alexa, Denis Zorin, and Daniele Panozzo.
\newblock Abc: A big cad model dataset for geometric deep learning.
\newblock In {\em The IEEE Conference on Computer Vision and Pattern Recognition (CVPR)}, June 2019.

\bibitem{lambourne2021brepnet}
Joseph~G Lambourne, Karl~DD Willis, Pradeep~Kumar Jayaraman, Aditya Sanghi, Peter Meltzer, and Hooman Shayani.
\newblock Brepnet: A topological message passing system for solid models.
\newblock In {\em Proceedings of the IEEE/CVF Conference on Computer Vision and Pattern Recognition}, pages 12773--12782, 2021.

\bibitem{Sketch2CAD_2020_li}
Changjian Li, Hao Pan, Adrien Bousseau, and Niloy~J. Mitra.
\newblock Sketch2cad: Sequential cad modeling by sketching in context.
\newblock {\em ACM Trans. Graph. (Proceedings of SIGGRAPH Asia 2020)}, 39(6):164:1--164:14, 2020.

\bibitem{GRASS_sig17li}
Jun Li, Kai Xu, Siddhartha Chaudhuri, Ersin Yumer, Hao Zhang, and Leonidas Guibas.
\newblock Grass: Generative recursive autoencoders for shape structures.
\newblock {\em ACM Transactions on Graphics (Proc. of SIGGRAPH 2017)}, 36(4):to appear, 2017.

\bibitem{deepmc_liao2018}
Yiyi Liao, Simon Donne, and Andreas Geiger.
\newblock Deep marching cubes: Learning explicit surface representations.
\newblock In {\em Proceedings of the IEEE Conference on Computer Vision and Pattern Recognition}, pages 2916--2925, 2018.

\bibitem{L_scher_2019}
Christoph Lüscher, Eugen Beck, Kazuki Irie, Markus Kitza, Wilfried Michel, Albert Zeyer, Ralf Schlüter, and Hermann Ney.
\newblock Rwth asr systems for librispeech: Hybrid vs attention.
\newblock {\em Interspeech 2019}, Sep 2019.

\bibitem{occnet_mescheder2019}
Lars Mescheder, Michael Oechsle, Michael Niemeyer, Sebastian Nowozin, and Andreas Geiger.
\newblock Occupancy networks: Learning 3d reconstruction in function space.
\newblock In {\em Proceedings of the IEEE/CVF Conference on Computer Vision and Pattern Recognition}, pages 4460--4470, 2019.

\bibitem{mo2019structurenet}
Kaichun Mo, Paul Guerrero, Li Yi, Hao Su, Peter Wonka, Niloy Mitra, and Leonidas~J Guibas.
\newblock Structurenet: Hierarchical graph networks for 3d shape generation.
\newblock 2019.

\bibitem{polygen_nash20}
Charlie Nash, Yaroslav Ganin, S.~M.~Ali Eslami, and Peter Battaglia.
\newblock {P}oly{G}en: An autoregressive generative model of 3{D} meshes.
\newblock In Hal~Daumé III and Aarti Singh, editors, {\em Proceedings of the 37th International Conference on Machine Learning}, volume 119 of {\em Proceedings of Machine Learning Research}, pages 7220--7229. PMLR, 13--18 Jul 2020.

\bibitem{para2021sketchgen}
Wamiq~Reyaz Para, Shariq~Farooq Bhat, Paul Guerrero, Tom Kelly, Niloy Mitra, Leonidas Guibas, and Peter Wonka.
\newblock Sketchgen: Generating constrained cad sketches.
\newblock {\em arXiv preprint arXiv:2106.02711}, 2021.

\bibitem{deepsdf_park2019}
Jeong~Joon Park, Peter Florence, Julian Straub, Richard Newcombe, and Steven Lovegrove.
\newblock Deepsdf: Learning continuous signed distance functions for shape representation.
\newblock In {\em Proceedings of the IEEE/CVF Conference on Computer Vision and Pattern Recognition}, pages 165--174, 2019.

\bibitem{imageTrans_parmar2018}
Niki Parmar, Ashish Vaswani, Jakob Uszkoreit, Lukasz Kaiser, Noam Shazeer, Alexander Ku, and Dustin Tran.
\newblock Image transformer.
\newblock In {\em International Conference on Machine Learning}, pages 4055--4064. PMLR, 2018.

\bibitem{pointnet2_qi2017}
Charles~R Qi, Li Yi, Hao Su, and Leonidas~J Guibas.
\newblock Pointnet++: Deep hierarchical feature learning on point sets in a metric space.
\newblock {\em arXiv preprint arXiv:1706.02413}, 2017.

\bibitem{PixeCNN_Salimans2017}
Tim Salimans, Andrej Karpathy, Xi Chen, and Diederik~P. Kingma.
\newblock Pixelcnn++: A pixelcnn implementation with discretized logistic mixture likelihood and other modifications.
\newblock In {\em ICLR}, 2017.

\bibitem{csgnet_sharma2018}
Gopal Sharma, Rishabh Goyal, Difan Liu, Evangelos Kalogerakis, and Subhransu Maji.
\newblock Csgnet: Neural shape parser for constructive solid geometry.
\newblock In {\em Proceedings of the IEEE Conference on Computer Vision and Pattern Recognition}, pages 5515--5523, 2018.

\bibitem{parsenet_sharma2020}
Gopal Sharma, Difan Liu, Subhransu Maji, Evangelos Kalogerakis, Siddhartha Chaudhuri, and Radom{\'\i}r M{\v{e}}ch.
\newblock Parsenet: A parametric surface fitting network for 3d point clouds.
\newblock In {\em European Conference on Computer Vision}, pages 261--276. Springer, 2020.

\bibitem{tian2019learning}
Yonglong Tian, Andrew Luo, Xingyuan Sun, Kevin Ellis, William~T Freeman, Joshua~B Tenenbaum, and Jiajun Wu.
\newblock Learning to infer and execute 3d shape programs.
\newblock {\em arXiv preprint arXiv:1901.02875}, 2019.

\bibitem{transformer_NIPS2017}
Ashish Vaswani, Noam Shazeer, Niki Parmar, Jakob Uszkoreit, Llion Jones, Aidan~N Gomez, \L~ukasz Kaiser, and Illia Polosukhin.
\newblock Attention is all you need.
\newblock In I. Guyon, U.~V. Luxburg, S. Bengio, H. Wallach, R. Fergus, S. Vishwanathan, and R. Garnett, editors, {\em Advances in Neural Information Processing Systems}, volume~30. Curran Associates, Inc., 2017.

\bibitem{walke2020learning}
Homer Walke, R~Kenny Jones, and Daniel Ritchie.
\newblock Learning to infer shape programs using latent execution self training.
\newblock {\em arXiv preprint arXiv:2011.13045}, 2020.

\bibitem{pixel2mesh_wang2018}
Nanyang Wang, Yinda Zhang, Zhuwen Li, Yanwei Fu, Wei Liu, and Yu-Gang Jiang.
\newblock Pixel2mesh: Generating 3d mesh models from single rgb images.
\newblock In {\em Proceedings of the European Conference on Computer Vision (ECCV)}, pages 52--67, 2018.

\bibitem{pienet_NEURIPS2020}
Xiaogang Wang, Yuelang Xu, Kai Xu, Andrea Tagliasacchi, Bin Zhou, Ali Mahdavi-Amiri, and Hao Zhang.
\newblock Pie-net: Parametric inference of point cloud edges.
\newblock In H. Larochelle, M. Ranzato, R. Hadsell, M.~F. Balcan, and H. Lin, editors, {\em Advances in Neural Information Processing Systems}, volume~33, pages 20167--20178. Curran Associates, Inc., 2020.

\bibitem{sceneformer_wang2020}
Xinpeng Wang, Chandan Yeshwanth, and Matthias Nie{\ss}ner.
\newblock Sceneformer: Indoor scene generation with transformers.
\newblock {\em arXiv preprint arXiv:2012.09793}, 2020.

\bibitem{edge_1985Weiler}
K. {Weiler}.
\newblock Edge-based data structures for solid modeling in curved-surface environments.
\newblock {\em IEEE Computer Graphics and Applications}, 5(1):21--40, 1985.

\bibitem{topologicalSF_Weiler1986}
Kevin Weiler.
\newblock Topological structures for geometric modeling.
\newblock 1986.

\bibitem{willis2021engineering}
Karl~DD Willis, Pradeep~Kumar Jayaraman, Joseph~G Lambourne, Hang Chu, and Yewen Pu.
\newblock Engineering sketch generation for computer-aided design.
\newblock In {\em Proceedings of the IEEE/CVF Conference on Computer Vision and Pattern Recognition}, pages 2105--2114, 2021.

\bibitem{fusion2020willis}
Karl D.~D. Willis, Yewen Pu, Jieliang Luo, Hang Chu, Tao Du, Joseph~G. Lambourne, Armando Solar-Lezama, and Wojciech Matusik.
\newblock Fusion 360 gallery: A dataset and environment for programmatic cad reconstruction.
\newblock {\em arXiv preprint arXiv:2010.02392}, 2020.

\bibitem{3dgan_wu2016}
Jiajun Wu, Chengkai Zhang, Tianfan Xue, Bill Freeman, and Josh Tenenbaum.
\newblock Learning a probabilistic latent space of object shapes via 3d generative-adversarial modeling.
\newblock In {\em Advances in Neural Information Processing Systems}, pages 82--90, 2016.

\bibitem{pqnet_wu2020}
Rundi Wu, Yixin Zhuang, Kai Xu, Hao Zhang, and Baoquan Chen.
\newblock Pq-net: A generative part seq2seq network for 3d shapes.
\newblock In {\em Proceedings of the IEEE/CVF Conference on Computer Vision and Pattern Recognition}, pages 829--838, 2020.

\bibitem{xu2021inferring}
Xianghao Xu, Wenzhe Peng, Chin-Yi Cheng, Karl~DD Willis, and Daniel Ritchie.
\newblock Inferring cad modeling sequences using zone graphs.
\newblock In {\em Proceedings of the IEEE/CVF Conference on Computer Vision and Pattern Recognition}, pages 6062--6070, 2021.

\bibitem{pointflow_yang2019}
Guandao Yang, Xun Huang, Zekun Hao, Ming-Yu Liu, Serge Belongie, and Bharath Hariharan.
\newblock Pointflow: 3d point cloud generation with continuous normalizing flows.
\newblock In {\em Proceedings of the IEEE/CVF International Conference on Computer Vision}, pages 4541--4550, 2019.

\bibitem{foldingnet_yang2018}
Yaoqing Yang, Chen Feng, Yiru Shen, and Dong Tian.
\newblock Foldingnet: Point cloud auto-encoder via deep grid deformation.
\newblock In {\em Proceedings of the IEEE Conference on Computer Vision and Pattern Recognition}, pages 206--215, 2018.

\end{thebibliography}
}

\clearpage
\setcounter{page}{1}
\appendix
\title{Supplementary Document \\ DeepCAD: A Deep Generative Network for Computer-Aided Design Models}
\maketitle

\section{CAD dataset}
\label{sec:supp_dataset}
\begin{figure*}[t!]
\begin{center}
\includegraphics[width=0.99\textwidth, height=0.45\textwidth]{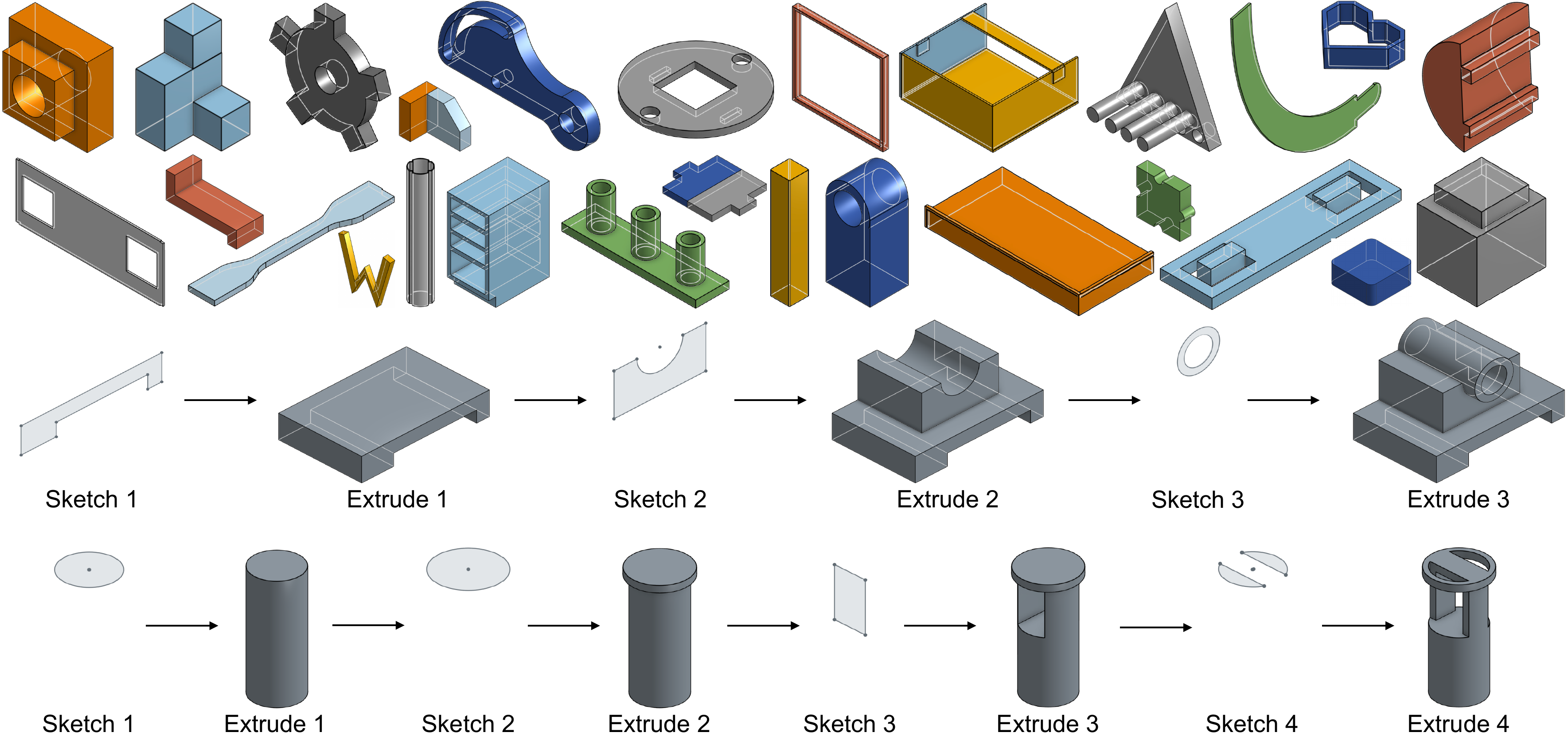}
\end{center}
\vspace{-4mm}
\caption{\textbf{Our CAD dataset}. Top two rows: examples of CAD models in our dataset. 
Bottom two rows: examples of CAD construction sequences.}
\label{fig:dataset_gallery}
\end{figure*}

We create our dataset by parsing the command sequences of CAD models in Onshape's online
repository. Some example models from our dataset are shown in~\figref{dataset_gallery}.
Unlike other 3D shape datasets which have specific categories (such as chairs and cars), 
this dataset are mostly user-created mechanical parts, and they have diverse shapes. 

Our dataset is derived from the ABC dataset~\cite{abc_2019_CVPR}, which contains several duplicate shapes. 
So for each shape in the test set, we find its nearest neighbor in the training set based on chamfer distance; we discard this test shape if the nearest distance is below a threshold.

\begin{figure}[h!]
    \centering
\includegraphics[width=0.99\linewidth]{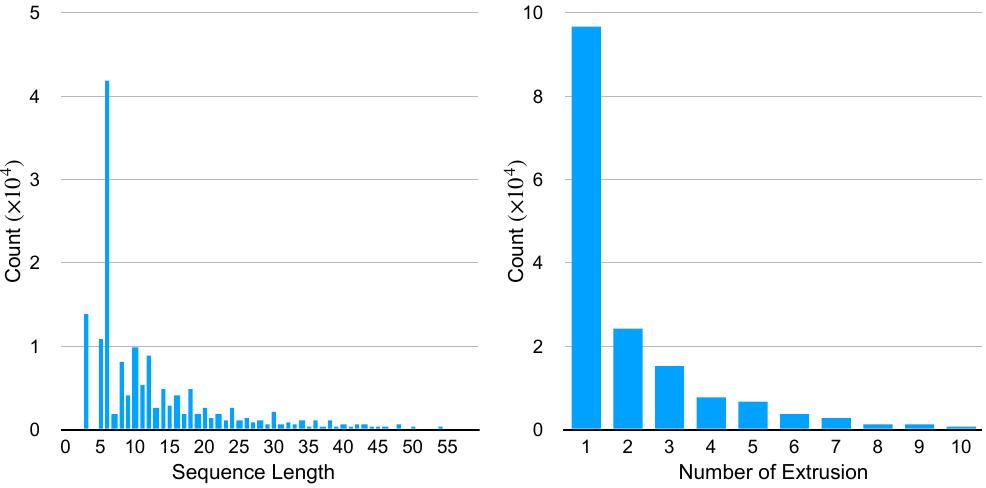}
\vspace{-0.5mm}
   \caption{Statistics of CAD training data in terms of command sequence length (left) and number of extrusions (right).}
\label{fig:dataset_count}
\vspace{-3mm}
\end{figure}

We further examine the data distribution in terms of CAD command sequence
length and the number of extrusions (see~\figref{dataset_count}).  Most CAD
command sequences are no longer than $40$ or use less than $8$ extrusions, as
these CAD models are all manually created by the user.  A similar command
length distribution is also reported in Fusion 360
Gallery~\cite{fusion2020willis}.

\section{Command Parameter Representation}\label{sec:supp_datarep}
Recall that our list of the full command parameters is $\bm{p}_i = [x, y, \alpha, f,
r,\theta, \phi, \gamma, p_x, p_y, p_z, s, e_1, e_2, b, u]$
in \tabref{data_rep}.  As described in~\secref{rep}, we 
normalize and quantize these parameters. 

First, we scale every CAD model within a $2\times 2\times 2$ cube
(without translation) such that all parameters stay bounded:
the sketch plane origin $(p_x, p_y, p_z)$ 
and the two-side extrusion distances $(e_1, e_2)$ range in $[-1, 1]$;  
the scale of associated sketch profile $s$ is within $[0, 2]$;  and the
sketch orientation $(\theta,\phi,\gamma)$ have the range $[-\pi,\pi]$.

Next, we normalize every sketch profile into a unit square such that its
starting point (i.e., the bottom-left point) locates at the center $(0.5,
0.5)$.  As a result, the curve's ending position $(x, y)$ and the radius $r$ of a circle
stay within $[0, 1]$.  The arc's sweeping angle $\alpha$ by definition is in $[0, 2\pi]$.

Afterwards, we quantize all continuous parameters into $256$ levels and express them using $8$-bit integers.

For discrete parameters, we directly use their values. 
The arc's counter-clockwise flag $f$ is a binary sign: $0$ indicates
clockwise arc and $1$ indicates counter-clockwise arc.  
The CSG operation type $b\in\{0, 1, 2, 3\}$ indicates \emph{new body}, \emph{join},
\emph{cut} and \emph{intersect}, respectively. Lastly, the extrusion type $u\in\{0,
1, 2\}$ indicates \emph{one-sided}, \emph{symmetric} and \emph{two-sided}, respectively.

\section{Network Architecture and Training Details}\label{sec:supp_arch}
\paragraph{Autoencoder.} 
Our Transformer-based encoder and decoder are both composed of four layers of
Transformer blocks, each with eight attention heads and a feed-forward
dimension of $512$.  We adopt standard layer normalization and a dropout rate
of $0.1$ for each Transformer block.

The last Transformer block in the decoder is followed by two separate linear
layers, one for predicting command type (with weights $W_1\in
\mathbb{R}^{256\times 6}$), and another for predicting command parameters (with weights
$W_2\in \mathbb{R}^{256\times 4096}$).  The output, a $4096$-dimensional vector,
from the second linear layer is further reshaped into a matrix of shape $16\times
256$, which indicates each of the total $16$ parameters.

\paraspace
\paragraph{Latent-GAN.} 
Section~\ref{sec:training} describes the use of latent-GAN technique on our
learned latent space for CAD generation.  
In our GAN model, the generator and discriminator are
both MLP networks, each with four hidden layers. Every hidden layer has a dimension of $512$.  The
input dimension (or the noise dimension) is $64$, and the output dimension is
$256$.  We use WGAN-gp strategy~\cite{wgan_pmlr-v70-arjovsky17a,wgangp_Gulrajani2017} to train the network:
the number of critic iterations is set to $5$ and the weight factor for gradient
penalty is set to $10$.  
The training lasts for $200,000$ iterations with a batch
size of $256$. In this process, Adam optimizer is used with a learning rate of $2\times 10^{-4}$
and $\beta_1=0.5$.

\section{Autoencoding CAD models}\label{sec:supp_ae}
\paragraph{Comparison methods.}
Here we describe in details the variants of our method used in~\secref{results_ae} for comparison.

\nameRel represents curve positions relative to the position of its previous
curve in the loop. 
As a result, the ending positions of a line and a arc and the center of a circle
differ from those in our method,
but the representation of other curve parameters (i.e., $\alpha, f, r$ in \tabref{data_rep}) stays 
the same. 

\nameTrans includes in the extrusion command the starting position ($s_x,
s_y$) of the loop, in addition to the origin of the sketch plane.  The origin
($p_x, p_y, p_z$) is in the world frame of reference; the loop's starting
position ($s_x, s_y$) is described in the local frame of the sketch plane.
In our proposed approach, however, we translate the sketch plane's origin to the loop's starting position. 
Thereby, there is no need to specify the parameters ($s_x, s_y$) explicitly.

\nameArc specifies an arc using its ending and middle positions.  
As a result, the representation of an arc becomes into $(x, y, m_x, m_y)$, where $(x,
y)$ indicates the ending position (as in our method), but $(m_x, m_y)$ is used to indicate the arc's middle point.

\nameRegr regresses all parameters of the CAD commands using the standard
mean-squared error in the loss function. 
The Cross-Entropy loss for discrete parameters (such as command types) stays the same as our proposed approach.
But in this variant, continuous parameters are not quantized, 
although they are still normalized into the range $[-1, 1]$
in order to balance the mean-squared errors introduced by different parameters.

\nameAug includes randomly composed CAD command
sequences in its training process. 
This is a way of data augmentation.
When we randomly choose a CAD model from the dataset during training, 
there is $50\%$ chance that the sampled CAD sequence will be mixed with another
randomly sampled CAD sequence.  The mixture of the two CAD command sequences is
done by randomly switching one or more pairs of sketch and extrusion (in their
commands).  CAD sequences that contain only one pair of sketch and extrusion
are not involved in this process.

\begin{table}[b]
\begin{center}
\begin{tabular}{lccc}
\toprule
Method & \makecell{mean \\ CD} & \makecell{trimmed mean \\ CD} & \makecell{median \\ CD}\\
\midrule
\nameAug    & \textbf{6.14} & \textbf{0.974} & \textbf{0.752} \\
$\mathtt{Ours}$ & 7.16 & 1.08 & 0.787 \\
\nameArc & 6.90 & 1.09 & 0.790 \\
\nameTrans  & 7.14 & 1.09 & 0.792 \\
\nameRel    & 9.24 & 1.38 & 0.863 \\ 
\nameRegr   &  12.61    & 3.87     & 2.14 \\
\bottomrule
\end{tabular}
\end{center}
\vspace{-1mm}
\caption{Mean, trimmed mean and median chamfer distances for shape autoencoding. Numerical values are multiplied by $10^3$. }
\label{tab:supp_cd}
\end{table}

\begin{figure}[t!]
\begin{center}
\includegraphics[width=0.99\linewidth]{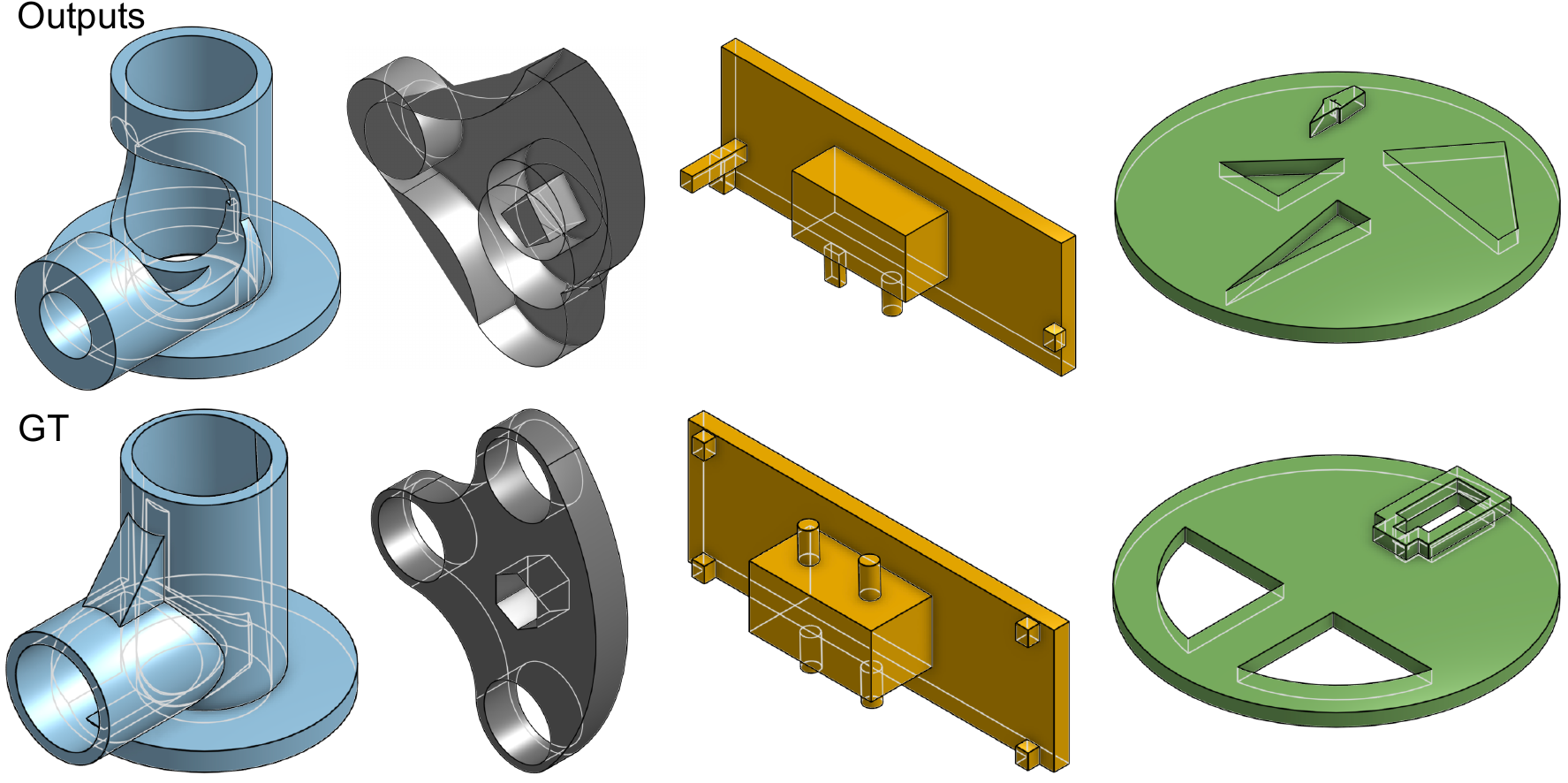}
\end{center}
\vspace{-4mm}
   \caption{Failure examples in shape autoencoding. Top: our reconstructed CAD outputs. Bottom: ground-truth CAD models. }
\label{fig:failures}
\end{figure}

\paraspace
\paragraph{Full statistics for CD scores.} 
In~\tabref{supp_cd}, we report the mean, trimmed mean, and median chamfer distance (CD) scores for our CAD autoencoding study.  
``Trimmed mean'' CD is computed by
removing $10\%$ largest and $10\%$ smallest scores.  
The mean CD scores are significantly higher than the trimmed mean and median CD scores. 
This is because the prediction of CAD sequence in some cases may be sensitive to small 
perturbations: a small change in command sequence may lead to a large change of shape
topology and may even invalidate the topology (e.g., the gray shape in~\figref{failures}). 
Those cases happen rarely, but when they happen,
the CD scores become significantly large. It is those outliers that make the mean CD scores much higher.

\begin{figure}[t]
\begin{center}
\includegraphics[width=0.99\linewidth]{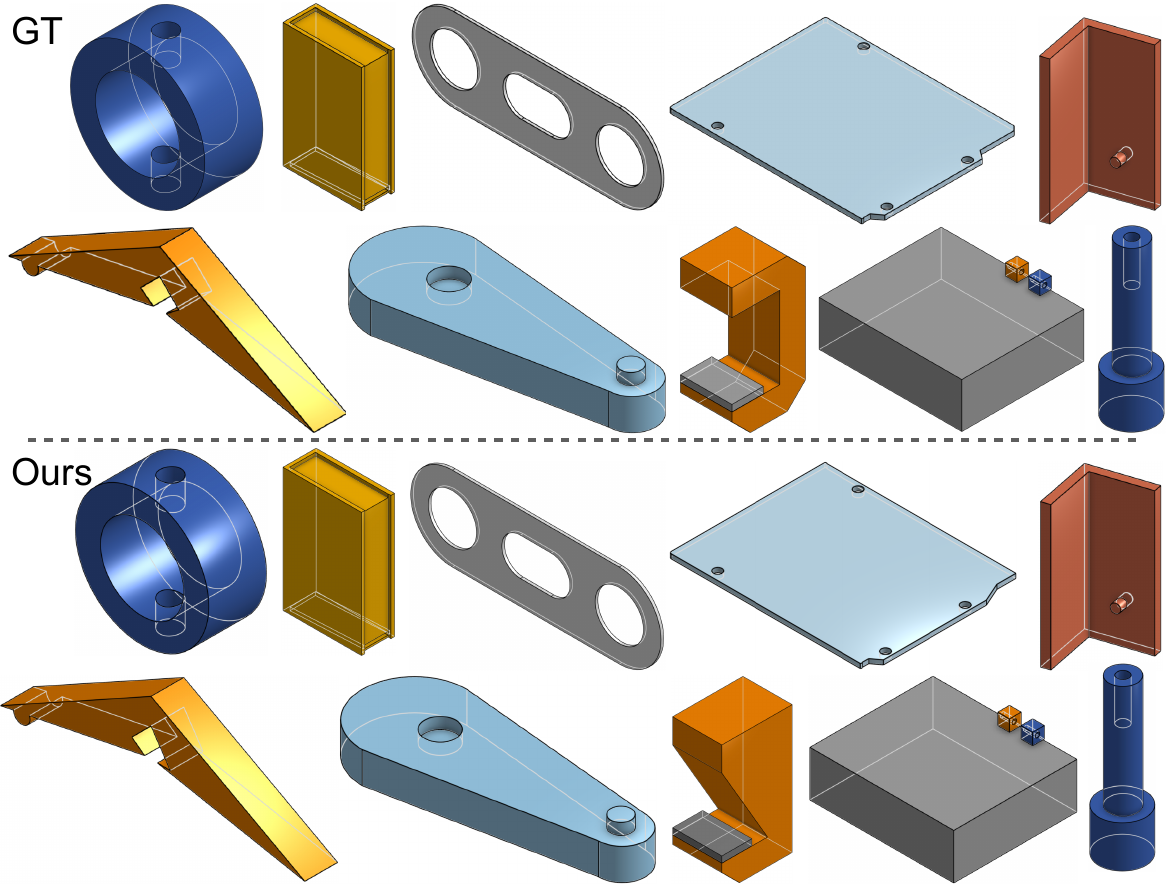}
\end{center}
\vspace{-4mm}
\caption{Shape autoencoding results on Fusion 360 Gallery~\cite{fusion2020willis} test data.
The autoencoder is trained using our own dataset.}
\label{fig:results_fusion}
\end{figure}

\begin{table}[t]
\begin{center}
\begin{tabular}{cccc}
\toprule
\makecell{$\accCmd\uparrow$} & \makecell{$\accParam\uparrow$} & \makecell{median \\ CD$\downarrow$} & \makecell{Invalid \\ Ratio$\downarrow$}\\
\midrule
97.90 & 96.45 & 0.796 & 1.62 \\
\bottomrule
\end{tabular}
\end{center}
\vspace{-3mm}
\caption{Quantitative evaluation for shape autoencoding on Fusion 360
    Gallery~\cite{fusion2020willis} test data. The model is only trained on our
proposed dataset. $\uparrow$: the higher the better, $\downarrow$: the lower
the better.}
\label{tab:results_fusion}
\end{table}

\begin{figure}[t!]
\begin{center}
\includegraphics[width=0.99\linewidth, height=0.45\linewidth]{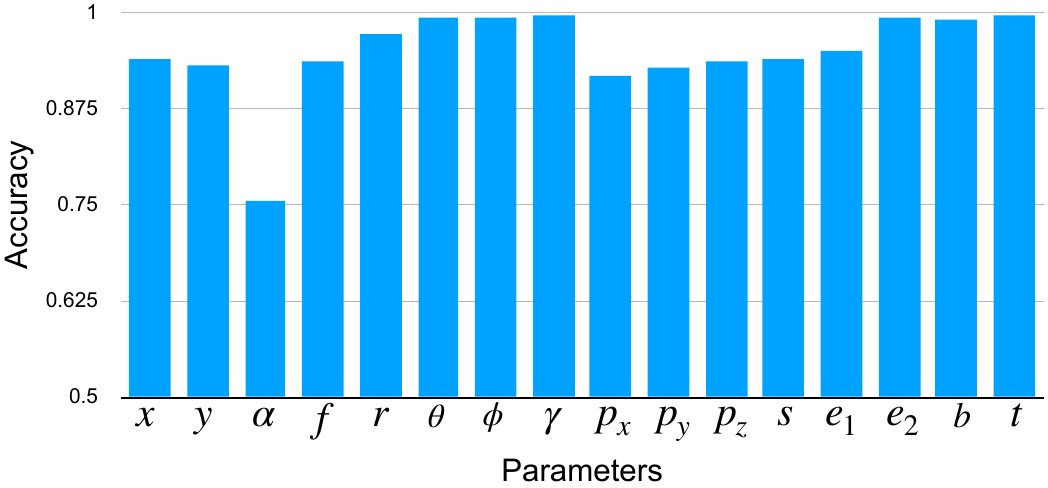}
\end{center}
\vspace{-4mm}
\caption{Accuracies for individual parameter types.}
\label{fig:acc_wrt_param}
\end{figure}

\paraspace
\paragraph{Accuracies of individual parameter types.}
We also examine the accuracies for individual types of parameters.
The accuracy is defined in \secref{results_ae}, and the results are shown in~\figref{acc_wrt_param}.
While all the parameter are treated equally in the loss function,
their accuracies have some differences.
Most notably, the recovery of arc's sweeping angle $\alpha$ has lower accuracy than other parameters.
By examining the dataset, we find that the values of sweeping angle $\alpha$
span over its value range (i.e., $[0, 2\pi]$) more evenly than other parameters, but the arc
command is much less frequently used than other commands.
Thus,  in comparison to other parameters, it is harder to learn the recovery of the arc sweeping angle.

\begin{figure}[t!]
\begin{center}
\includegraphics[width=0.99\linewidth, height=0.45\linewidth]{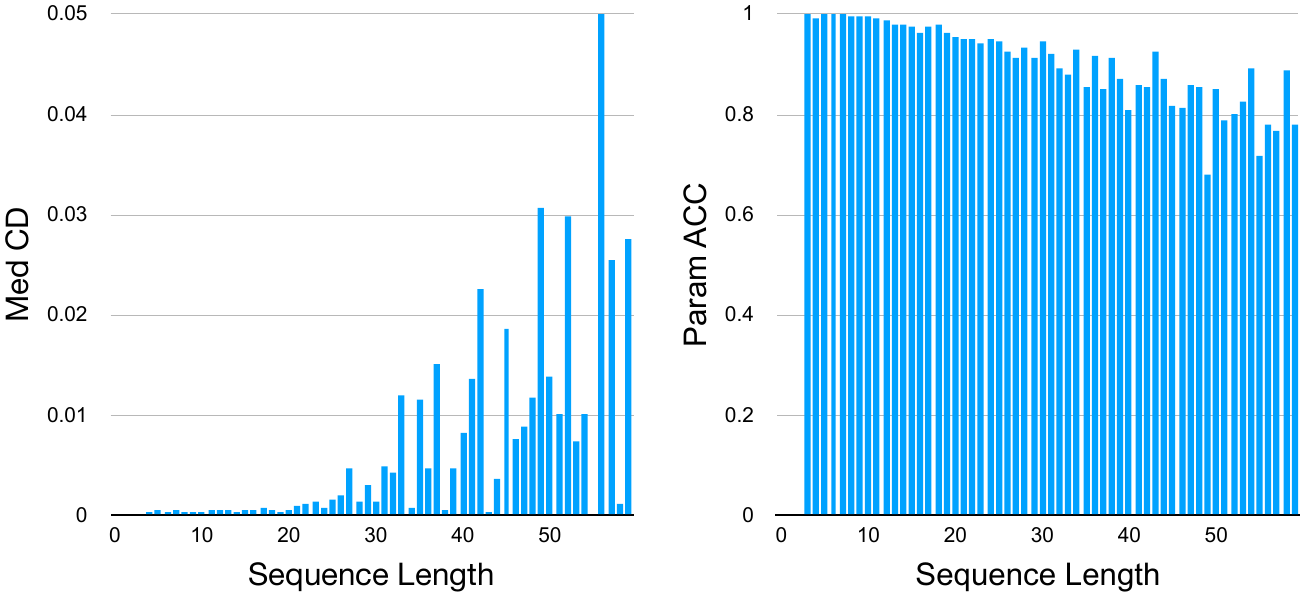}
\end{center}
\vspace{-4mm}
   \caption{Quantitative metrics for shape autoencoding \wrt CAD sequence
   length. Left: median chamfer distance (the lower the better). Right:
   parameter accuracy (the higher the better).}
\label{fig:metrics_wrt_len}
\end{figure}

\begin{table}[h]
\begin{center}
\resizebox{\linewidth}{!}{
\begin{tabular}{lcccc}
\toprule
Method & \makecell{$\accCmd\uparrow$} & \makecell{$\accParam\uparrow$} & \makecell{median \\ CD$\downarrow$} & \makecell{Invalid \\ Ratio$\downarrow$}\\
\midrule
$\mathtt{Ours}$ & 85.95 & 74.22 & 10.30 & 12.08 \\
$\mathtt{Ours\text{-}noise}$ & 84.65 & 74.23 & 10.44 & 13.82 \\
\bottomrule
\end{tabular}}
\end{center}
\vspace{-2.5mm}
\caption{Quantitative results for CAD reconstruction from point clouds. 
$\mathtt{Ours\text{-}noise}$ corresponds to noisy inputs (uniform noise in $[-0.02, 0.02]$ along normal direction). We use the same metrics as in autoencoding task; $\accCmd$ and $\accParam$ are both multiplied by $100\%$, and CD is multiplied by $10^3$.}
\label{tab:results_pc2cad}
\vspace{-3mm}
\end{table}

\begin{figure*}[t!]
    \centering
\includegraphics[width=0.99\textwidth, height=0.38\linewidth]{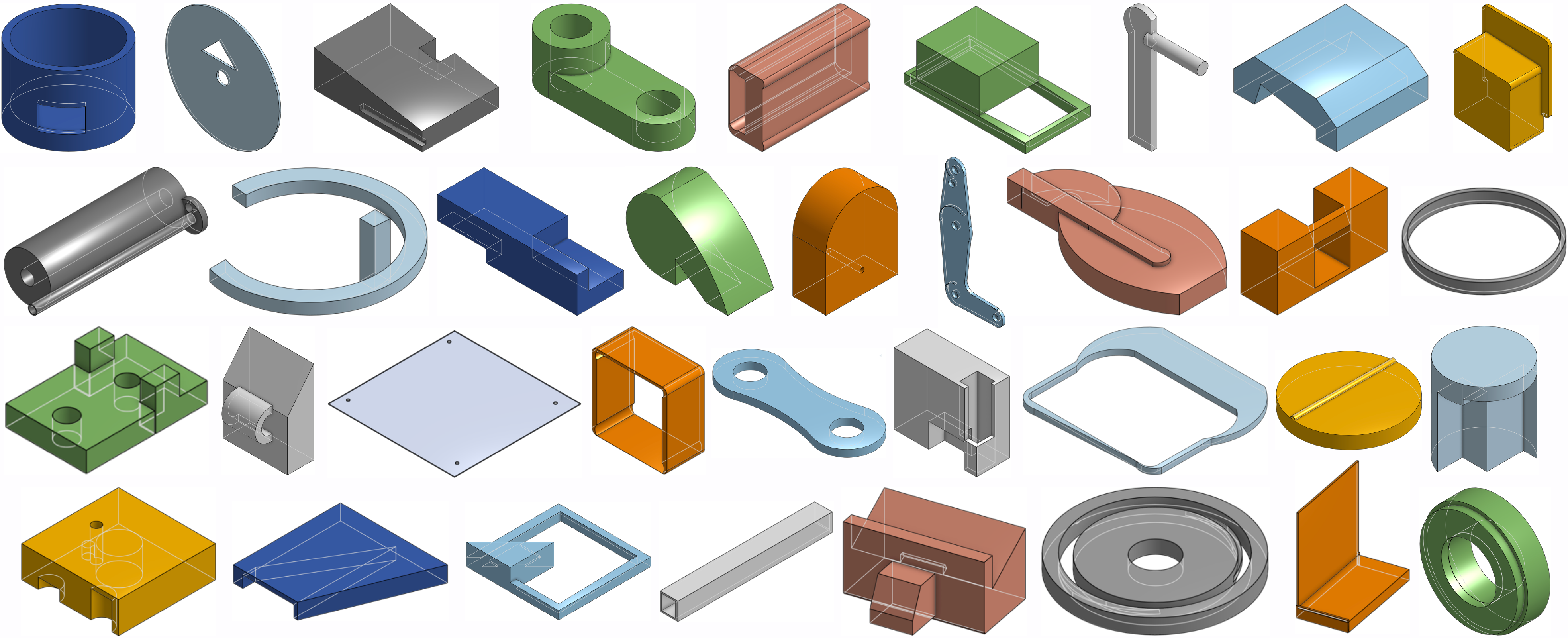}
   \caption{A gallery of our generated CAD models.}\label{fig:supp_gen}
\end{figure*}

\section{Generalization on Fusion 360 Gallery~\cite{fusion2020willis}}
\label{sec:supp_fusion}
To validate the generalization ability of our autoencoder, we perform a
cross-dataset test.  
In particular, for shape autoencoding tasks,  
we take the model trained on our proposed
dataset and evaluate it using a different dataset provided by Fusion 360
Gallery~\cite{fusion2020willis}. 
These two datasets are constructed
from different sources: ours is based on models from Onshape repository, whereas theirs is created
from designs in Autodesk Fusion 360.
The qualitative and quantitative results are
shown in~\figref{results_fusion} and~\tabref{results_fusion}, respectively,
showing that our trained model performs well on shape distributions that are different
from the training dataset.

\section{Failure Cases}\label{sec:supp_failure}
Not every CAD command sequence is valid.
Our method 
is more likely to produce invalid CAD commands when the command length becomes long. 
Figure~\ref{fig:failures} shows a few failed results.
The produced gray shape has invalid topology, and the yellow shape suffers from
misplacement of small sketches.
Figure~\ref{fig:metrics_wrt_len} plots the median CD scores and
the parameter accuracies with respect to CAD command sequence length.  
The difficulties for generating long-sequence CAD models are twofold.
As the CAD sequence becomes longer, it is harder to ensure valid topology.
Meanwhile, as shown in~\figref{dataset_count}, the data
distribution in terms of the sequence length has a long tail;
the dataset provides much more short sequences than long sequences.  This data imbalance 
may cause the network model to bias toward short sequences.

\section{Metrics for Shape Generation}\label{sec:supp_metrics}
We follow the three metrics used in~\cite{pmlr-v80-achlioptas18a} to evaluate
the quality of our shape generation. 
In~\cite{pmlr-v80-achlioptas18a}, these metrics are motivated 
for evaluating the point-cloud generation.
Therefore, for computing these metrics for CAD models, we first convert them into point clouds.
Then, these metrics are defined by comparing 
a set of reference shapes $\mathcal{S}$ with
a set of generated shapes $\mathcal{G}$. 

\textbf{Coverage (COV)} measures the diversity of generated shapes by computing
the fraction of shapes in the reference set $\mathcal{S}$ that are matched by at least
one shape in the generated set $\mathcal{G}$.  
Formally, COV is defined as
\begin{equation}
    \text{COV}(\mathcal{S}, \mathcal{G})=\frac{|\{\arg\min_{Y\in \mathcal{S}} d^{\text{CD}}(X, Y)| X\in \mathcal{G}\}|}{|\mathcal{S}|},
\end{equation}
where $d^{\text{CD}}(X, Y)$ denote the chamfer distance between two point clouds $X$ and $Y$.

\textbf{Minimum matching distance (MMD)} measures the fidelity of generated shapes.
For each shape in the reference set $\mathcal{S}$, the chamfer distance to its nearest
neighbor in the generated set $\mathcal{G}$ is computed.  MMD is defined as the average
over all the nearest distances:
\begin{equation}
    \text{MMD}(\mathcal{S}, \mathcal{G})=\frac{1}{|\mathcal{S}|}\underset{Y\in \mathcal{S}}{\sum}\underset{X\in \mathcal{G}}\min d^{\text{CD}}(X,Y).
\end{equation}

\textbf{Jensen-Shannon Divergence (JSD)} is a statistical distance
metric between two data distributions. Here, it measures the similarity between the reference set $\mathcal{S}$ and the
generated set $\mathcal{G}$ by computing the marginal point distributions:
\begin{equation}
    \text{JSD}(P_{\mathcal{S}}, P_{\mathcal{G}})=\frac{1}{2}D_{\text{KL}}(P_\mathcal{S}||M)+\frac{1}{2}D_{\text{KL}}(P_{\mathcal{G}}||M),
\end{equation}
where $M=\frac{1}{2}(P_{\mathcal{S}}+P_{\mathcal{G}})$ and $D_{\text{KL}}$ is the standard KL-divergence.
$P_{\mathcal{S}}$ and $P_{\mathcal{G}}$ are marginal distributions of points in the reference and
generated sets, approximated by discretizing the space into $28^3$ voxel grids
and assigning each point from the point cloud to one of them.

Since our full test set is
relatively large, we randomly sample a reference set of $1000$ shapes and
generate $3000$ shapes using our method to compute the metric scores. 
To reduce the sampling bias,
we repeat this evaluation process for three times and report the average scores.

\end{document}